\documentclass[10pt,twocolumn,letterpaper]{article}

\usepackage{cvpr}              

\usepackage[dvipsnames]{xcolor}

\definecolor{turquoise}{cmyk}{0.65,0,0.1,0.3}
\definecolor{purple}{rgb}{0.65,0,0.65}
\definecolor{dark_green}{rgb}{0, 0.5, 0}
\definecolor{orange}{rgb}{0.8, 0.6, 0.2}
\definecolor{red}{rgb}{0.8, 0.2, 0.2}
\definecolor{darkred}{rgb}{0.6, 0.1, 0.05}
\definecolor{blueish}{rgb}{0.0, 0.3, .6}
\definecolor{light_gray}{rgb}{0.7, 0.7, .7}
\definecolor{pink}{rgb}{0.9, 0, 0.6}
\definecolor{greyblue}{rgb}{0.25, 0.25, 1}
\definecolor{teal}{rgb}{0.0, 0.4, 0.4}

\newcommand{\ky}[1]{{\color{dark_green}#1}}

\def \customparskip {.2em}
\renewcommand{\paragraph}[1]{\vspace{\customparskip}\noindent\textbf{#1}}

\definecolor{cvprblue}{rgb}{0.21,0.49,0.74}
\usepackage[pagebackref,breaklinks,colorlinks,citecolor=cvprblue]
{hyperref}
\usepackage{booktabs,tabularx}
\usepackage[ruled,linesnumbered]{algorithm2e}
\usepackage{mathtools}
\usepackage{amsmath}
\usepackage{multirow}
\usepackage{appendix}
\DeclareMathOperator*{\argmin}{arg\,min}
\usepackage{graphicx}
\newcommand*{\img}[1]{%
    \raisebox{-.3\baselineskip}{%
        \includegraphics[
        height=\baselineskip,
        width=\baselineskip,
        keepaspectratio,
        ]{#1}%
    }%
}
\newcolumntype{Y}{>{\hsize=3\hsize}X}

\def\paperID{2640} 
\def\confName{CVPR}
\def\confYear{2024}

\title{Prompting Hard or Hardly Prompting: \\ Prompt Inversion for Text-to-Image Diffusion Models
}

\author{Shweta Mahajan$^{1,2}$ ~~~~~~~~~~~~ Tanzila Rahman$^{1,2}$ ~~~~~~~~~~~~ Kwang Moo Yi$^{1}$ ~~~~~~~~~~~~ Leonid Sigal$^{1,2}$ \\
$^1$University of British Columbia ~~~~~~~~~~~~~~ $^2$Vector Institute for AI\\
~~~~~~~~ Vancouver, BC, Canada ~~~~~~~~~~~~~~~~~~~~~ Toronto, ON, Canada\\
{\tt\small \{smahaj06, trahman8, kmyi, lsigal\}@cs.ubc.ca}
}

\begin{document}
\maketitle
\begin{abstract}
The quality of the prompts provided to text-to-image diffusion models determines how faithful the generated content is to the user's intent, often requiring `prompt engineering'.
To harness visual concepts from target images without prompt engineering, current approaches largely rely on embedding inversion by optimizing and then mapping them to pseudo-tokens.
However, working with such high-dimensional vector representations is challenging because they lack semantics and interpretability, and only allow simple vector operations when using them.
Instead, this work focuses on inverting the diffusion model to obtain interpretable language prompts directly. 
The challenge of doing this lies in the fact that the resulting optimization problem is fundamentally discrete and the space of prompts is exponentially large; 
this makes using standard optimization techniques, such as stochastic gradient descent, difficult.
To this end, we utilize a delayed projection scheme to optimize for prompts representative of the vocabulary space in the model.
Further, we leverage the findings that different timesteps of the diffusion process cater to different levels of detail in an image. 
The later, noisy, timesteps of the forward diffusion process correspond to the semantic information, and therefore, prompt inversion in this range provides tokens representative of the image semantics. 
We show that our approach can identify semantically interpretable and meaningful prompts for a target image which can be used to synthesize diverse images with similar content.
We further illustrate the application of the optimized prompts in evolutionary image generation and concept removal.
\end{abstract}    
\section{Introduction}
\label{sec:intro}
\begin{figure}
    \centering
    \includegraphics[width=\columnwidth]{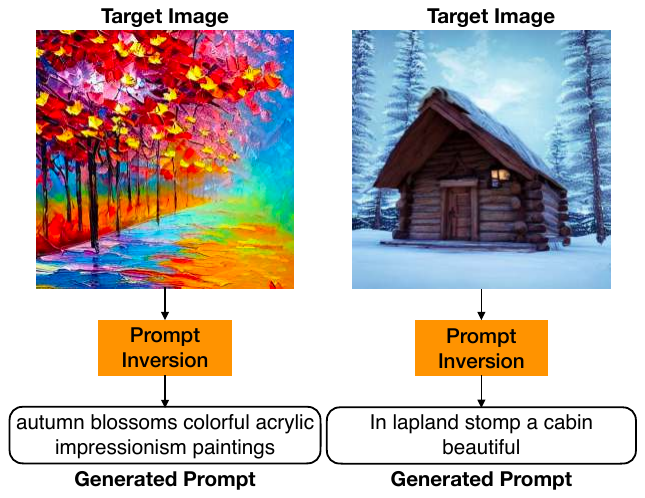}
    \caption{{\bf Illustration of Hard Prompt Inversion.} Target images (top) along with inverted prompts (bottom) obtained using proposed \emph{PH2P} approach; note the exhibited prompt semantics.}
    \label{fig:intro:teaser}
    \vspace{-15 pt}
\end{figure}

Text-to-image conditional diffusion models \cite{rombach2022high,SahariaCSLWDGLA22,nichol2021glide} are trained on an enormous amount of image-text data and have transformed the domain of generative learning in computer vision.
These models demonstrate exceptional generative capabilities and provide an opportunity for creative image generation with customized concepts \cite{RahmanLRTMS23,ramesh2021zero,JainXA23,PanZT23,singer2022make}. 
The quality and the regularity of the content produced by diffusion models are, however,  subject to the quality of the input prompts; which in turn depends on the training data \cite{ZhouYLL22,GuSZ20}.
For pre-trained diffusion models, identifying and formulating the prompts that produce the intended visual content is challenging in the large-scale data regime.

For example, to generate the painting shown in \cref{fig:intro:teaser}, one would require an understanding of art styles such as \emph{acrylic} or \emph{impressionism}.
These concepts, though encoded in the diffusion model, might not be accessible without specialized domain knowledge or without access to the model's training vocabulary.
Additionally, in many downstream applications of diffusion models such as image composition \cite{WenHPME23,KumariZ0SZ23,LiuLDTT22} or segmentation \cite{abs-2303-11681} target concepts are available solely as images and require the discovery of the appropriate prompt as an intermediate step.
This drives a relevant research question: \emph{What is a likely prompt that would generate visual contents of the target image?}
Discovering, or optimizing, such prompts would greatly simplify creative and editing processes; particularly for novice users.

Prompts for the visual content of interest are predominantly hand-crafted through laborious trial and error requiring human intervention and expertise \cite{abs-2307-12980,NiHFZ23,HertzMTAPC23}. 
Inspired by the inversion techniques that search semantics of the target images in the latent space of generative adversarial networks (GANs) \cite{BojanowskiJLS18,CreswellB19a,ZhuKSE16,GuSZ20}, recent work has instead resorted to the automated discovery of the target visual concepts through inversion of diffusion models \cite{KumariZ0SZ23,GalAAPBCC23,abs-2303-09522,RuizLJPRA23}. 
With text-guided synthesis at its core, text-conditioned diffusion models apply inversion to the conditioning representations.
Specifically, new pseudo-tokens are introduced in the model vocabulary and the corresponding embeddings are optimized for the target image(s). 
These embeddings, however, have been found to display less or no correlation to the original model vocabulary \cite{RuizLJPRA23}.
Moreover, a single vector representation encodes varied concepts for a collection of images, thereby restricting the readability and generalization capabilities of the learned concepts. 

The aforementioned limitations can be addressed by inverting the diffusion model to yield the text tokens \ie hard prompts within the model's vocabulary.\footnote{`Hard' here also refers to the fact that these tokens are discrete choices, and not `soft' weightings of the training vocabulary, which result in continuous vectors in the text embedding space.} 
Optimization of the hard prompts involves a discrete optimization procedure with re-projection of the learned embedding onto the nearest neighbor in the pre-specified vocabulary \cite{WenHPME23}.  
Hard prompt inversion in diffusion models is particularly cumbersome owing to the complexity of the generative model with a two-stage pipeline consisting of the conditioning network (CLIP text encoder; \cite{RadfordKHRGASAM21}) and the generative component (a U-Net \cite{RonnebergerFB15}) making the flow of gradients to the input layer difficult (vanishing gradients).
Secondly, the diffusion trajectory has high variance given the stochastic components in the generative model \cite{SongME21}.
As a result, it can be observed (see Sec.~\ref{sec:experiments}), that standard SGD-based optimization with projection leads to poor performance in practice.

\begin{figure*}[t!]
\begin{subfigure}{1.1\columnwidth}
    \centering
    \includegraphics[width=\linewidth]{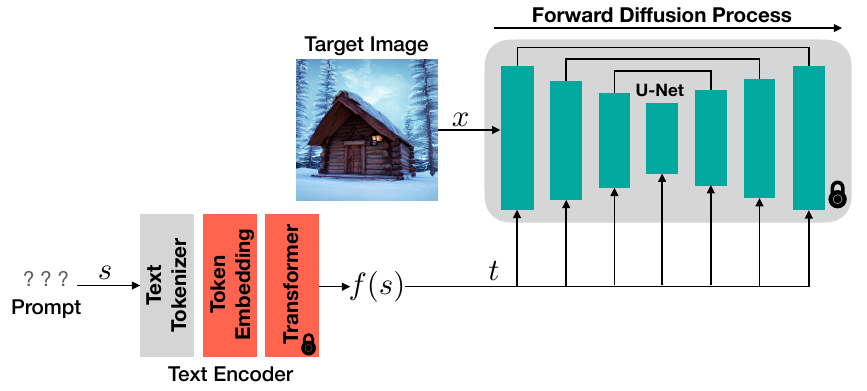}
    \caption{Overview of prompt optimization with Latent Diffusion Model.}
    \label{fig:archi}
\end{subfigure}
\hfill
\begin{subfigure}{\columnwidth}
    \centering
    \includegraphics[width=0.8\linewidth]{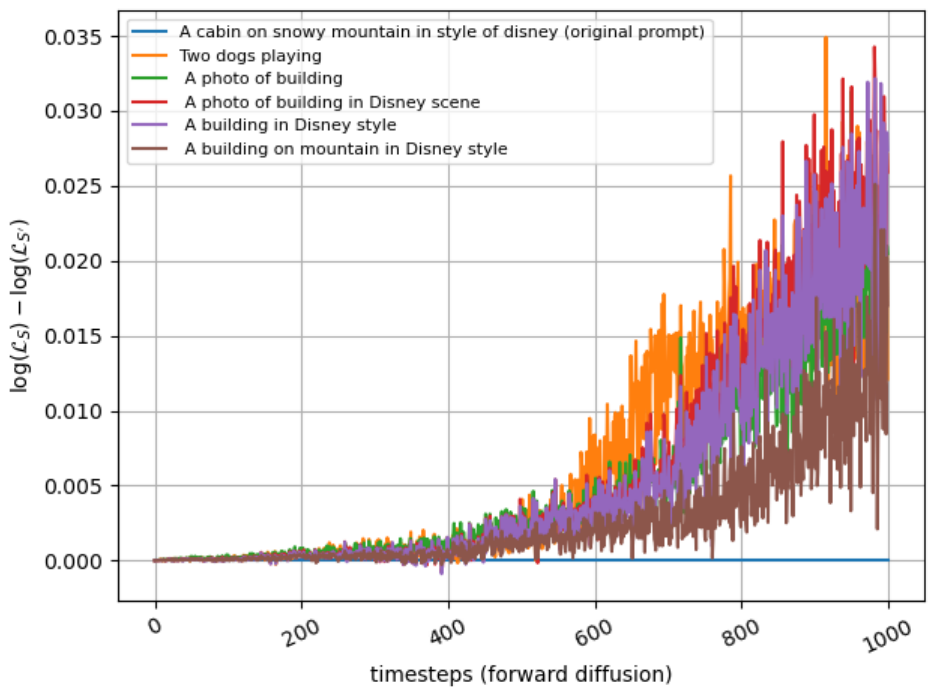}
    \caption{Effect of the conditioning on diffusion objective at different timesteps.}
    \label{fig:loss_comp}
\end{subfigure}
\caption{\textbf{Prompt Inversion in Diffusion Models.
} \emph{Left}: Overview of the text-to-image diffusion model with a CLIP text-encoder. \emph{Right}: Analysis of the sensitivity of different timesteps to conditioning information for a given target image illustrated on the left.}
\vspace{-5 mm}
\end{figure*}

We overcome these challenges for hard prompt inversion 
and make the following significant contributions:
\begin{itemize}
\item 
We study the effect of prompt conditioning at different timesteps of the diffusion process
(\cref{fig:loss_comp}).
We observe that noisy, later steps, have greater sensitivity to prompt conditioning.
\item  Based on our findings, we propose a \emph{Prompting Hard or Hardly Prompting (PH2P)} inversion procedure where the diffusion loss is applied to the sub-range of the timesteps, thereby reducing the high variance of the optimization process.
Prompt inversion in the conditioning-sensitive range also provides with better flow of gradients. 
We employ quasi-newton L-BFGS \cite{shanno1970conditioning} based reprojection techniques to learn discrete tokens. 
This provides us with a refined framework for sensitive multi-token optimization.  
\item We empirically validate that our approach yields semantically meaningful prompts that can synthesize accurate yet diverse images for a target visual concept. 
The prompts obtained with our proposed inversion technique are interpretable and applicable to a broad set of tasks, such as evolutionary image generation and concept removal in images through negative visual concepts.
\end{itemize}

\section{Related Work}
\label{sec:related_work}

\vspace{-\customparskip}
\paragraph{Text-guided image synthesis.} Extension of deep generative models to conditional, controllable generation is well-founded in generative adversarial networks (GANs)~\cite{GoodfellowPMXWOCB14,ReedAYLSL16,tseng2020retrievegan,abs-1904-01480}, variational autoencoders (VAEs)~\cite{KingmaW13,ZhangGMO19,zhu2019dm} and normalizing flows~\cite{DinhKB14,DinhSB17,BhattacharyyaMF20}.
Early work for text-to-image synthesis built upon GANs ~\cite{ReedAYLSL16} and more recently on the normalizing flow-based priors \cite{MahajanG020} to align the image-text distributions.
High quality and realistic image synthesis is now possible with models such as  
DALL-E \cite{ramesh2021zero} and Cogview \cite{ding2021cogview} that build upon the more expressive neural architectures such as transformers \cite{vaswani2017attention} and discrete variational autoencoders (VQ-VAE) \cite{razavi2019generating}.

High-fidelity image generation has been revolutionized with diffusion models  \cite{ho2020denoising} and has gained traction for controllable generation \cite{nichol2021glide,rombach2022high,gu2022vector,abs-2302-05543}.
Nichol \etal \cite{nichol2021glide} use CLIP \cite{RadfordKHRGASAM21} guidance to replace class labels and classifier-free guidance for conditional generation.
Saharia \etal instead of training the text encoder on paired image-text data, utilize a pre-trained language model as a text-encoder \cite{DevlinCLT19}.
Recent approaches \cite{rombach2022high,gu2022vector} apply diffusion models to the low-dimensional representations of the input thereby reducing the complexity of the generative component.
These low-dimensional diffusion models trained on large-scale datasets have become a standard backbone for the current generative modeling tasks such as visual art generation \cite{JainXA23,PanZT23} or multi-frame generation in videos or stories \cite{singer2022make,RahmanLRTMS23}.
The quality of the synthesized content in these models is influenced by the guiding text. 
It is important to find the correct text that yields desired visual concepts of interest and therefore, in this work, we develop our approach with latent diffusion models as working backbone.

\vspace{0.05in}
\paragraph{Prompting in diffusion models.} The natural language descriptions are referred to as discrete prompts or hard prompts and have a significant impact on image generation in diffusion models \cite{abs-2307-12980}.
To this end, various approaches have explored diverse prompting techniques based on retrieval and captioning to enhance the quality of the prompts \cite{abs-2303-11681,NiHFZ23,HeS0XZTBQ23} and consequently image synthesis.
Wu \etal \cite{abs-2303-11681} introduce sub-classes in the prompts based on the main class in images.
Ni \etal \cite{NiHFZ23} specify the name of the target object to be generated as input to a general purpose transformer (GPT) \cite{BrownMRSKDNSSAA20}.
Hertz \etal \cite{HertzMTAPC23} modify words in the text and inject the cross-attention maps of the pixels corresponding to target styles.
Our work, on the other hand, aims to generate the prompts directly from the text-prior within the diffusion model without any auxiliary task or model and without human intervention.

\paragraph{Inversion in diffusion models.}
Inspired by the optimization methods for inversion in GANs \cite{BojanowskiJLS18,CreswellB19a,ZhuKSE16,GuSZ20}, optimization-based approaches have been developed for diffusion models. 
In order to synthesize concepts with novel scenes, in diffusion models, the few approaches directly perform inversion in the image space \cite{ChoiKJGY21,DhariwalN21,SongME21,MokadyHAPC23}.
Song \etal and  Mokady \etal \cite{MokadyHAPC23} approaches search for the initial noise that reconstructs the image \cite{SongME21}.
Instead of working directly with the image features, recent approaches benefit from the controllability in the textual embedding space \cite{RuizLJPRA23,GalAAPBCC23,abs-2303-09522}.
At a high-level, these methods invert visual concepts in the embedding space of the text and encode new features as new tokens in the vocabulary. 
The DreamBooth approach of \cite{RuizLJPRA23} fine-tunes the diffusion model to learn a new visual featues and Kumari \etal update the parameters of the cross-attention layers. 
Textual inversion \cite{GalAAPBCC23} inverts the concepts from a set of reference images in the textual embedding space by introducing a new ``pseudo-word" and adding a new representation to the model vocabulary.
Voynov \etal \cite{abs-2303-09522} introduce a new token for each of the layers of the U-Net in the diffusion model.
The approaches provide better editing freedom to the users compared to approaches that modify the image space. 
However, the inverted features are not interpretable in the space of text and therefore, have limited generalization. 
Moreover, the recovered embeddings can be redundant and the image-level features may be present in the model vocabulary.
Recent, unpublished as of this submission, work of Wen \etal \cite{WenHPME23} attempts to optimize hard prompts for CLIP \cite{RadfordKHRGASAM21} with image-text similarity matching. 
Similar in spirit, to avoid biases, we propose hard prompt optimization that leverages text-conditioned diffusion model and its loss directly. 

\begin{figure*}
    \centering
    \includegraphics[width=\linewidth]{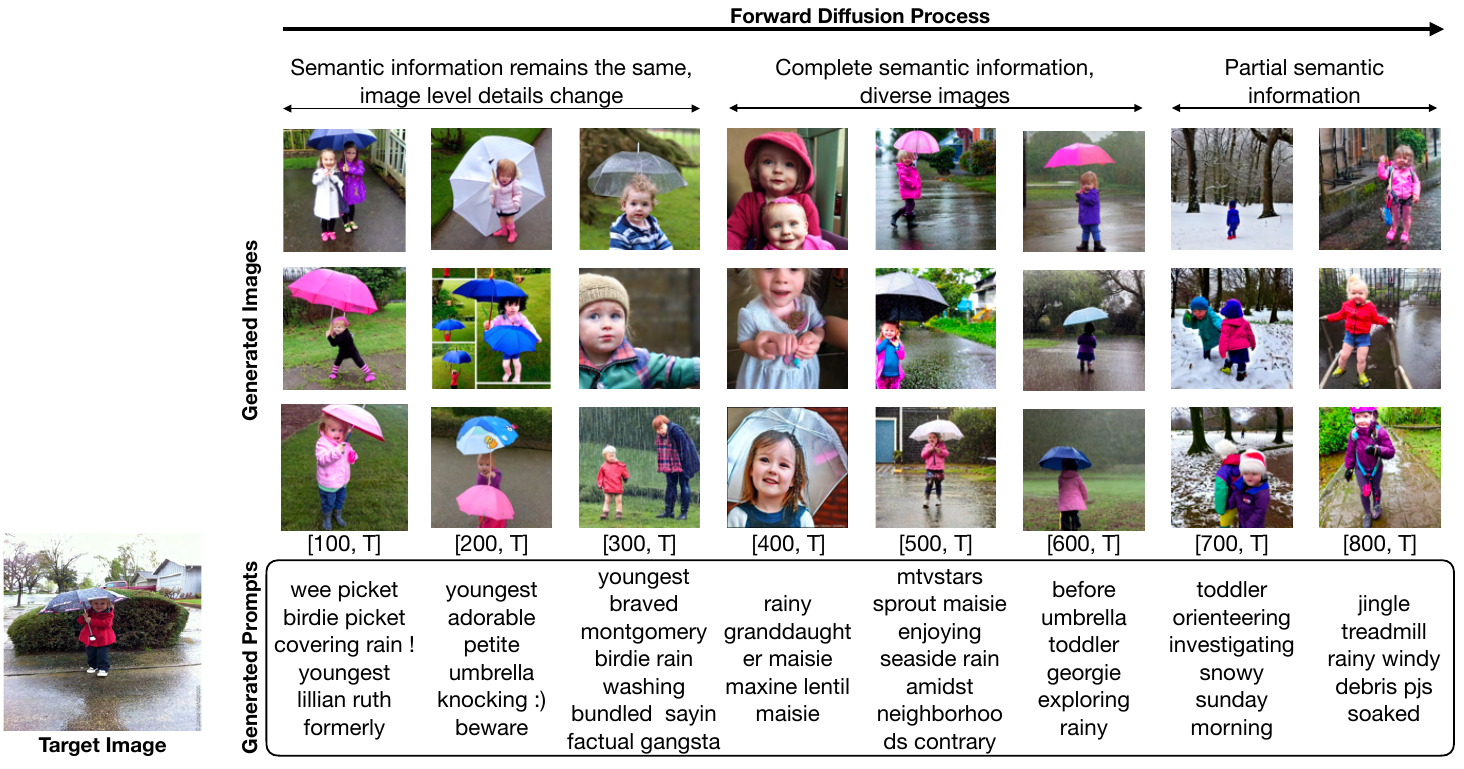}
    \caption{{\bf Effects of Timesteps on Prompt Inversion.} Prompts generated for the different sections of timesteps in text-to-image diffusion models with maximum noise at timestep $T$. The images generated show that the semantic information is completely recovered for the range starting from the middle timesteps. Once the semantic information appears in the images, only image-level details change when considering the larger range starting from the early steps of the forward diffusion process.}
    \label{fig:timestep_analysis}
\vspace{-5mm}
\end{figure*}

\section{Prompt Inversion for Diffusion}
While prior work for textual inversion  \cite{GalAAPBCC23,abs-2303-09522} is confined to the embedding space of the conditioning variable, in this work, we explicitly aim to find the text tokens that under the parameterized diffusion model
would yield the target image, \ie, \emph{what prompt would likely generate the desired image?}
We formalize our approach under the constraint that we do not have access to any additional auxiliary image-text similarity model and have access only to the pre-trained conditional diffusion model along with the text encoder used for conditioning.

\subsection{Conditional Diffusion Model Backbone}
We first provide an overview of the model components: the diffusion model with U-Net and the conditioning network, that we consider in the prompt inversion.

\paragraph{Latent diffusion model.} 
We consider the Latent Diffusion Model (LDM) \cite{rombach2022high} as the class of generative model for which we will recover the prompt tokens likely to generate the target image.
In LDM, the diffusion process is applied to a lower-dimensional spatial representation $\mathbf{x}$ of the input image $\mathbf{I}$.
An encoder $E(\cdot)$ maps the input image $\mathbf{I} \in \mathbb{R}^{H \times W \times 3}$ to a latent representation $\mathbf{x} \in \mathbb{R}^{h \times w \times c}$, downsampling the image to a lower spatial dimension.
The decoder $D(\cdot)$ maps $\mathbf{x}$ back to the original image space.
In a conditional setting, the diffusion model is applied to the representation $\mathbf{x}$, where time-conditioned U-Net \cite{RonnebergerFB15} $\epsilon_\theta(\mathbf{x}_t,t)$ is employed to model the diffusion process.
The diffusion process is applied over $T$ timesteps where noise is gradually added to $\mathbf{x}$ to generate a series of noisy representations $\mathbf{x}_{{1}:{T}}$.

For a conditional generative model, in addition to the timesteps, the U-Net is conditioned on $f(\mathbf{S})$, where conditioning input $\mathbf{S}$ is encoded with the mapping $f(\cdot)$.   
The objective of the conditional latent diffusion model is, 
\begin{align}
\mathcal{L}_{LDM} \coloneqq \mathbb{E}_{t,\mathbf{x},\epsilon}\left[\|\epsilon - \epsilon_\theta(\mathbf{x}_t,t,f(\mathbf{S}))\|_2^2\right].
\label{eq:LDM:obj}
\end{align}

During training, given the image $\mathbf{x}$ and the conditioning input \eg a prompt $\mathbf{S}$, a timestep is randomly chosen from $\{1,\ldots, T\}$ at each learning step and the model parameters $\theta$ that minimize \cref{eq:LDM:obj} are estimated.

\paragraph{Embedding space for optimization.} 
The text encoder $f(.)$ takes the input prompt $\mathbf{S}$, which is input to the latent diffusion model.
As shown in \cref{fig:archi}, the text tokenizer converts the prompt strings into tokens, the indices in a pre-specified vocabulary of size $|V|$.
Each  token is then mapped to its corresponding embedding vector $\mathbf{e}_i \in \mathbf{E}$ where $i \in \{1,\ldots,|V|\}$ in the embedding space $\mathcal{E}$.
The embedding vectors of the tokens in an input prompt are combined with the positional embeddings which are input to the transformer layer to get the encoded representation $f(\mathbf{S})$.

Prior work on textual inversion in the embedding space $\mathcal{E}$ ~\cite{GalAAPBCC23,abs-2303-09522} aims to learn new concepts for a set of images by introducing placeholder strings that are associated with a newly learned embedding vector. 
This entails extending the existing vocabulary with new concepts. 
These new concepts have, however, been found to be un-interpretable when read as natural language because the learned embeddings are typically far away from the embeddings of the original vocabulary.
In our approach, on the other hand,  given the target image, we aim to optimize the embedding vectors ${\mathbf{e}}_i$, $i \in \{1,\ldots, L\}$ from the existing vocabulary for a prompt with maximum length $L$.
This yields prompts that are human-readable.
Concurrent work in this direction optimizes the hard prompts with CLIP similarity \cite{WenHPME23} of the image and text encoding from the image and text CLIP encoders \cite{RadfordKHRGASAM21}.
In contrast, our approach is agnostic to the choice of encoders and depends entirely on the latent diffusion backbone utilized to perform prompt inversion. 


\subsection{Prompting Hard or Hardly Prompting} 
Given a target image $\mathbf{I}$ and the latent diffusion model parameterized by $\epsilon_\theta$, we optimize for the tokens in the existing vocabulary of the conditioning text encoder (CLIP text encoding) that best represent the visual content of the image. 
To learn the optimal tokens $\mathbf{S}^{*}$ that are likely to generate the target image, we use the following formulation as the optimization problem:
\begin{align}
\mathbf{S}^{*}=\argmin_{\mathbf{S}}\mathbb{E}_{t,\mathbf{x},\epsilon}[{\|\epsilon-\epsilon_\theta(\mathbf{x}_t,t,f(\mathbf{S}))\|_2^2}].
\label{eq:opt_obj}
\vspace{-5mm}
\end{align}
We aim to recover the tokens $\mathbf{S}^{*}$ which would satisfy the vocabulary of the conditional diffusion model.
Note that we keep the parameters of the text encoder and the diffusion model frozen.
Our approach, therefore, takes into account the language prior introduced in the generative model.

A naive approach to solving \cref{eq:opt_obj} would be to optimize for all diffusion time steps, with the typical SGD optimizer.
This, however, does not work in practice as we will show in \cref{sec:experiments}.
We thus introduce two key insights:
(1) we focus on time steps that actually matter;
(2) we keep our solutions \emph{strictly} within the vocabulary space that the model was trained with through projected gradient descent.

\paragraph{Timesteps for semantic information.}
To enable efficient and effective optimization, we investigate the impact conditioning has within a pre-trained diffusion model.
\Cref{fig:loss_comp} shows the diffusion loss of different prompts for a given image at different timesteps of the forward diffusion process.
We observe that the conditioning signal for the diffusion model is stronger at the later, more noisy timesteps of the diffusion process.
Conversely, the values of the diffusion loss are similar in the initial timesteps irrespective of the prompt used for the given image.

More specifically, consider the image (\cref{fig:archi}), which was generated with stable diffusion using the prompt ``A cabin on a snowy mountain in the style of Disney''. 
In \cref{fig:loss_comp} a random prompt ``Two dogs playing'' has an objective value very close to that of the original prompt in the initial timesteps up to $\sim 400$. 
The values of the objective diverge only for the later timesteps.
Moreover, when we provide random prompts with more information representative of the content of the image as well as the original prompt, the difference between the objective values of the random and the original prompts decreases. 
This hints that the high-level semantic information is encoded in the later timesteps while the low-level details are present in the initial timesteps of the forward diffusion process. 

Furthermore, in \cref{fig:timestep_analysis}, we show the effect of optimizing a prompt from different timestep ranges.
We observe that as we increase the range of timesteps to be optimized from more noise $t=T$ to less noise  $T=0$, the semantic information gradually increases up to the middle range of timesteps $\sim 500$. 
For the larger ranges including the initial timesteps of the forward diffusion, no new semantic details are recovered from the target image.

We use these key observations in our optimization refined \emph{prompting hard or hardly prompting (PH2P)} algorithm (\cref{algo:samp}) where we limit the range of $t$ to the later timesteps $\geq 500$. 
Optimizing only for the later timesteps has two-fold advantage. 
First, since we are optimizing for the initial layers of a very deep neural network, the gradients are very small. 
By optimizing for larger values of loss, we get better gradients and therefore, a better direction for descent in the high-dimensional space. 
Secondly, we empirically observe that the tokens in the generated prompts, when optimized for the earlier timesteps, do not add any new conditioning information for the prompt optimization and may contain special characters; see~\cref{fig:timestep_analysis}. 
Thus, optimizing in the noisy range of the diffusion process yields tokens that are more representative of the content.

\begin{algorithm}[t]
\SetNoFillComment
Input: Diffusion model parameters: $\theta$, Target image: $\mathbf{x}=E(\mathbf{I})$, Initial prompt: $\mathbf{S}$, Prompt embedding: $\mathbf{\hat{e}}$,
Timesteps: $[t_a, T]$; Learning rate: $\lambda$, Optimization steps: $N$\\
 \For{$i\gets 1 $ \KwTo $N$}{
 \tcc{Projection on feasible set}
        $\tilde{\mathbf{e}} =\text{Proj}_\mathbf{E}(\hat{\mathbf{e}}) $ \\
 \tcc{Select diffusion timestep}
        $t = \text{random}([t_a,T])$\\
  \tcc{Apply L-BFGS to \cref{eq:opt_obj}}
        $g=\text{LBFGS}_{\tilde{\mathbf{e}}}(\mathcal{L}_{LDM}(\mathbf{x}_t,\theta,t,f(\tilde{\mathbf{e}}))$\\
        $\hat{\mathbf{e}} = \hat{\mathbf{e}}-\lambda g$
        }
\tcc{Delayed projection}
return $\text{Proj}_\mathbf{E}(\hat{\mathbf{e}})$   

 \caption{PH2P Prompt Inversion}
 \label{algo:samp}
\end{algorithm}

\paragraph{Projected gradient descent for meaningful prompts.}
Following our observations, we choose the starting timestep $t_a < T$ and optimize \cref{eq:opt_obj} with $t \in [t_a, T]$ using projected gradient descent with delayed projections 
\cite{stich2020error,li2021delayed,courbariaux2015binaryconnect} in our PH2P prompt inversion for text-to-image diffusion models as shown in \cref{algo:samp}.
Starting with a random prompt, the loss is computed with respect to the embeddings in the feasible set \ie the embeddings representing the vocabulary.
During optimization, the variable $\mathbf{\hat{e}}$ is updated without the projection. 
This improves the efficiency of the descent.
Concurrent work builds upon a similar theory of delayed projected descent and applies the projection for CLIP similarity matching \cite{WenHPME23} with standard gradient descent.
However, when using a diffusion model, image encoders are not accessible and hence one cannot use projection with the image-text similarity loss.
In our work, at each iteration, the update is computed with respect to the projections of the updated embeddings using the L-BFGS \cite{shanno1970conditioning} algorithm.  
The variable updates performed in this manner provide for better convergence compared to projected descent with Adam optimizer; see~\cref{table:ablations}.
\section{Experiments}
\label{sec:experiments}
\begin{table*}
\scriptsize
\centering
  \begin{tabularx}{\linewidth}{@{}Xcccccc@{}}
  \toprule
    Method &  \multicolumn{3}{c}{COCO} &  \multicolumn{3}{c}{LSUN}\\
    \cmidrule(lr){2-4} \cmidrule(lr){5-7}
    & CLIP Similarity($\uparrow$) & LPIPS Similarity ($\downarrow$) & LPIPS Diversity ($\uparrow$)& CLIP Similarity($\uparrow$) & LPIPS Similarity ($\downarrow$) & LPIPS Diversity ($\uparrow$)\\
    \midrule
    PEZ \cite{WenHPME23} & 0.72 & 0.477 &0.417&0.70&0.480&0.420\\
    PH2P(Ours) & \bfseries 0.77 & \bfseries 0.462 & \bfseries 0.435 &\bfseries 0.77& \bfseries 0.463& \bfseries 0.422 \\
     \bottomrule
  \end{tabularx}
  \vspace{-2mm}
  \caption{{\bf Quantitative Evaluation.} Evaluation of the quality of the images generated with the prompts from inversion.}
  \label{table:coco_sun_quant}
\end{table*}
\begin{table*}[h]
 \centering
 \scriptsize
\begin{tabularx}{\textwidth}{@{}*{7}{m{0.12\textwidth}}@{}}
\toprule
Target Image & \multicolumn{3}{l}{PEZ \cite{WenHPME23}} & \multicolumn{3}{l}{PH2P (Ours)}\\
\cmidrule(lr){2-4} \cmidrule(lr){5-7}
  \raisebox{-18pt} {\includegraphics[height=0.12\textwidth, width = 0.12\textwidth]{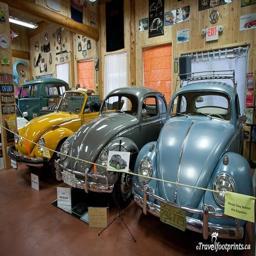}} &
 \raisebox{-18pt} {\includegraphics[height=0.12\textwidth, width = 0.12\textwidth]{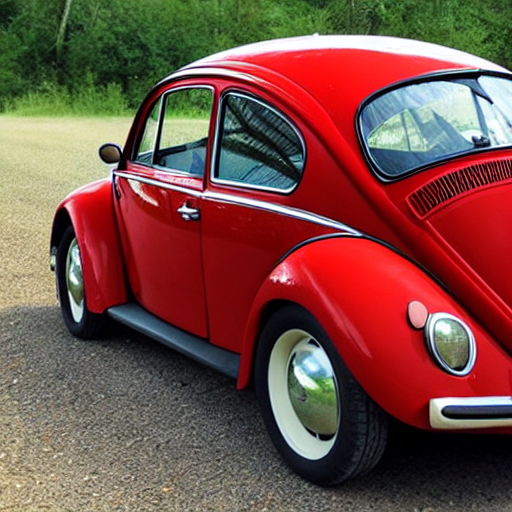}} &
\raisebox{-18pt}{\includegraphics[height=0.12\textwidth, width = 0.12\textwidth]{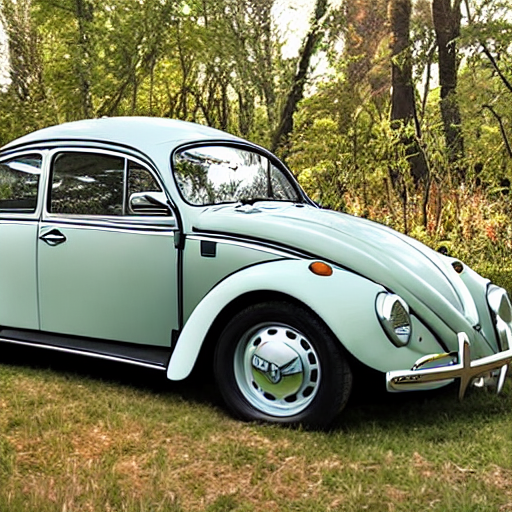}} &
\raisebox{-18pt}{\includegraphics[height=0.12\textwidth, width = 0.12\textwidth]{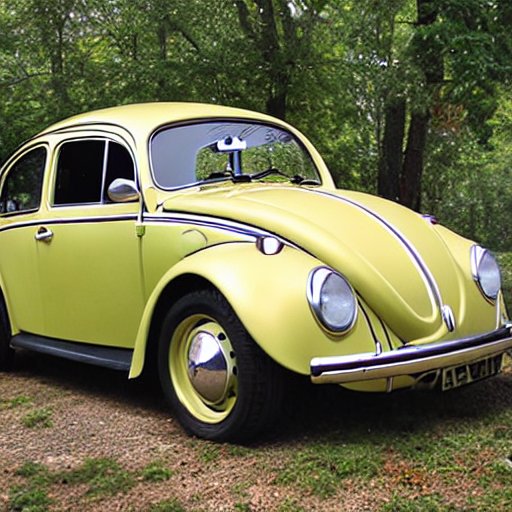}} &
\raisebox{-18pt}{\includegraphics[height=0.12\textwidth, width = 0.12\textwidth]{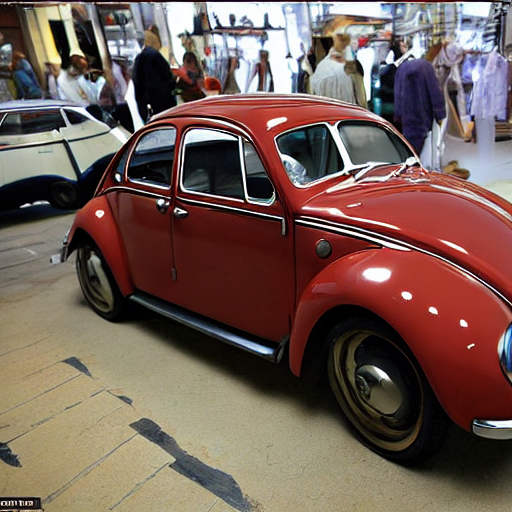}} &
\raisebox{-18pt}{\includegraphics[height=0.12\textwidth, width = 0.12\textwidth]{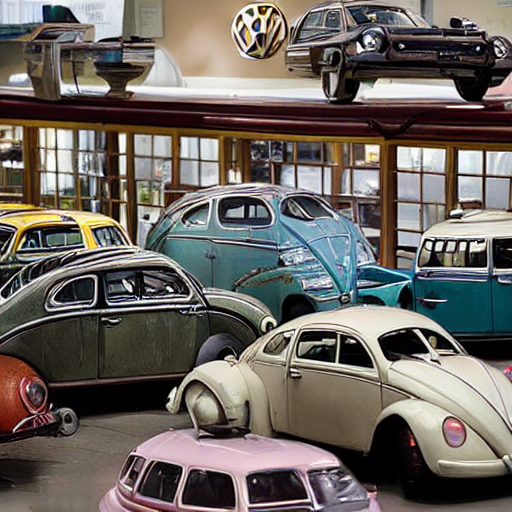}}&
\raisebox{-18pt}{\includegraphics[height=0.12\textwidth, width = 0.12\textwidth]{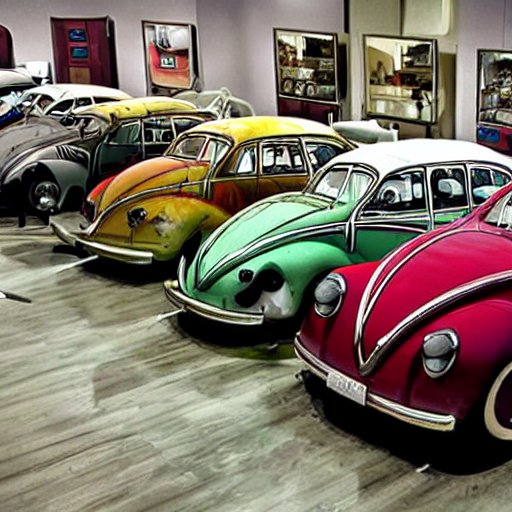}}\\
& \multicolumn{3}{m{0.36\textwidth+3\tabcolsep+\arrayrulewidth\relax}}{visited retro vw ptic unpopular beetle isn ldnont evie "@ oudiscoverydestinations dyk ote avalley} & \multicolumn{3}{m{0.36\textwidth+3\tabcolsep+\arrayrulewidth\relax}}{pressed shops converted volkswagen rat headlights seized rescued danube smithsonian museum dose tycoon outdated}\\
\midrule
   \raisebox{-18pt}{\includegraphics[height=0.12\textwidth, width = 0.12\textwidth]{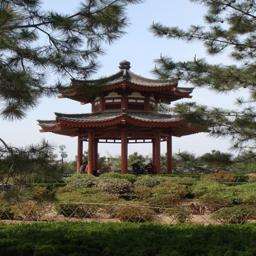}}&
 \raisebox{-18pt}{\includegraphics[height=0.12\textwidth, width = 0.12\textwidth]{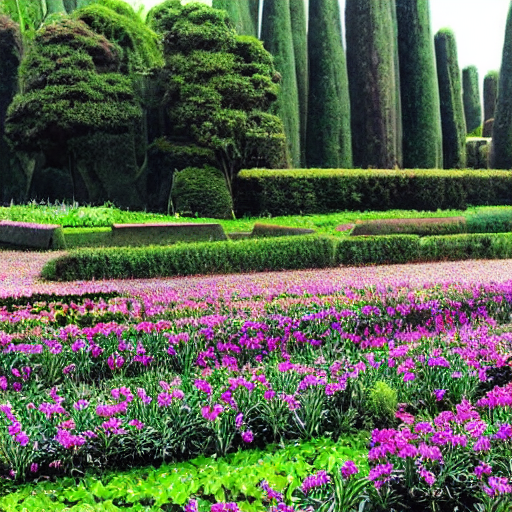}} &
 \raisebox{-18pt}{\includegraphics[height=0.12\textwidth, width = 0.12\textwidth]{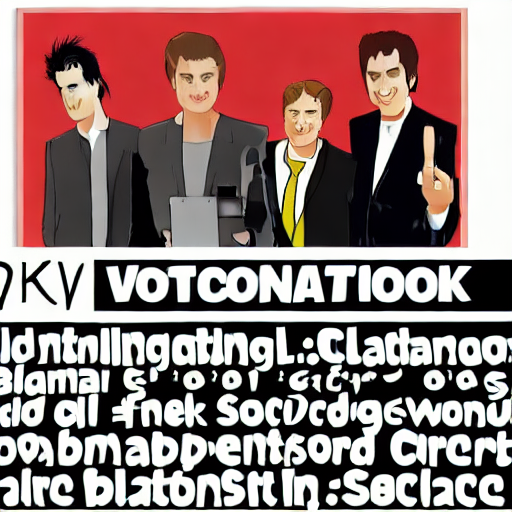}} &
 \raisebox{-18pt}{\includegraphics[height=0.12\textwidth, width = 0.12\textwidth]{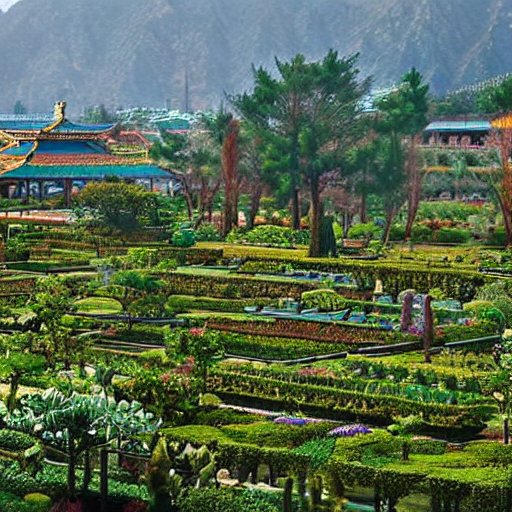}} &
 \raisebox{-18pt}{\includegraphics[height=2.0cm,width=2.0cm]{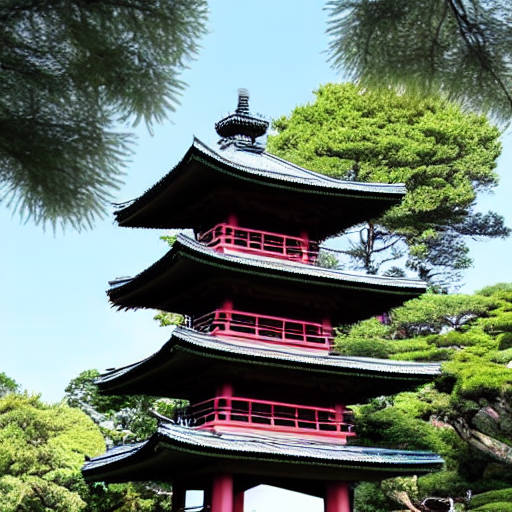}} &
 \raisebox{-18pt}{\includegraphics[height=0.12\textwidth, width = 0.12\textwidth]{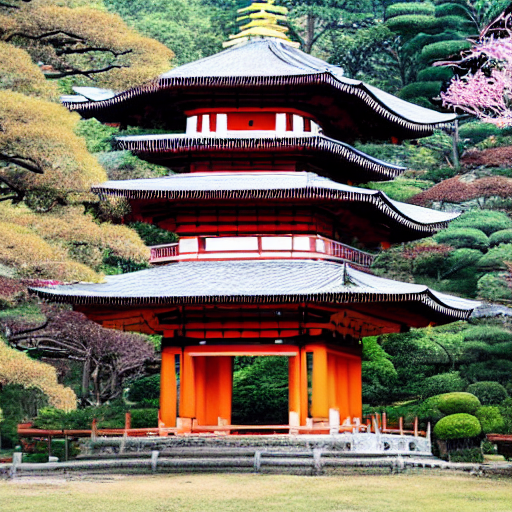}}&
 \raisebox{-18pt}{\includegraphics[height=0.12\textwidth, width = 0.12\textwidth]{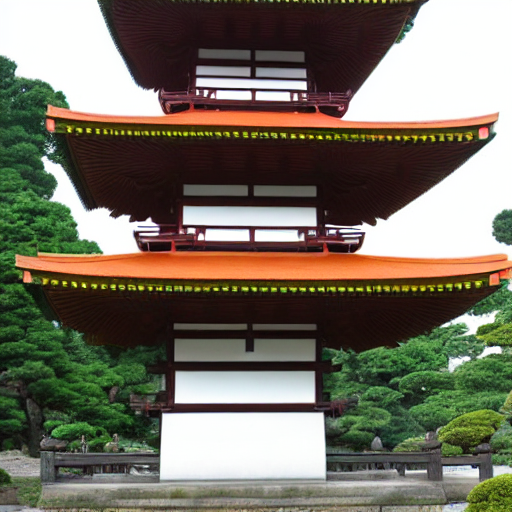}}\\
 &  \multicolumn{3}{m{0.36\textwidth+3\tabcolsep+\arrayrulewidth\relax}}{}
 
 & 
 \multicolumn{3}{m{0.36\textwidth+3\tabcolsep+\arrayrulewidth\relax}}{pavilion pagoda depicted japanese resources \newline bearing hilltop bearing mainland tang}\\
 \midrule
   \raisebox{-18pt}{\includegraphics[height=0.12\textwidth, width = 0.12\textwidth]{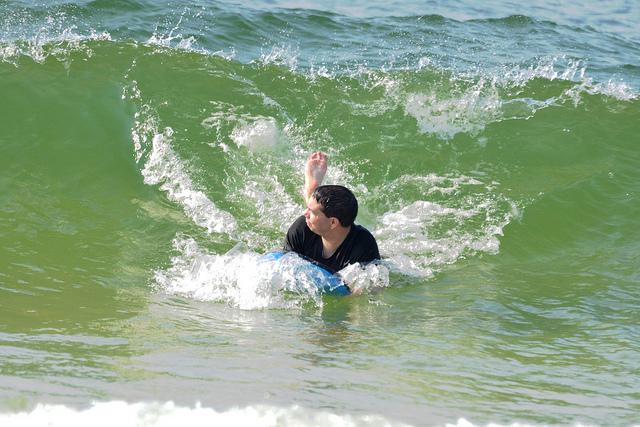}}&
 \raisebox{-18pt}{\includegraphics[height=0.12\textwidth, width = 0.12\textwidth]{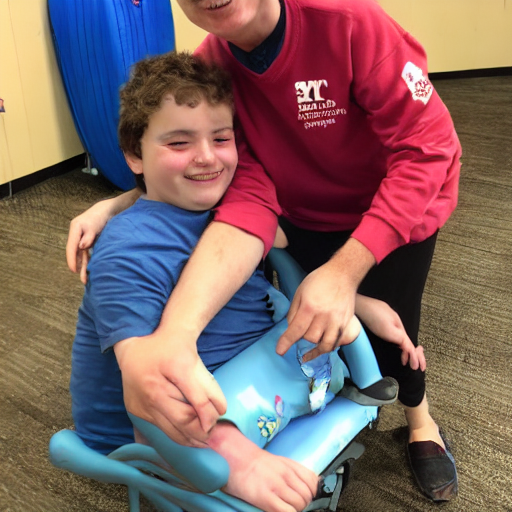}} &
 \raisebox{-18pt}{\includegraphics[height=0.12\textwidth, width = 0.12\textwidth]{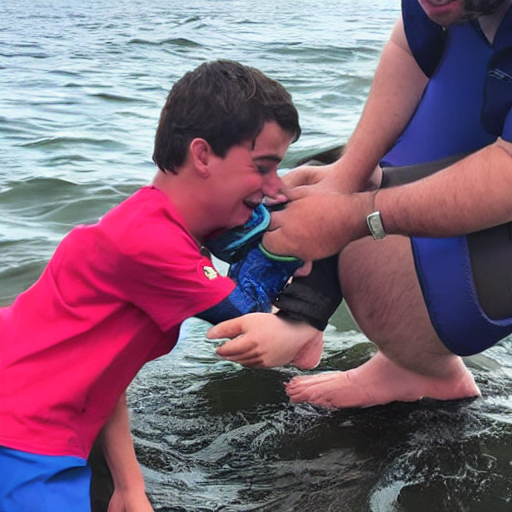}} &
 \raisebox{-18pt}{\includegraphics[height=0.12\textwidth, width = 0.12\textwidth]{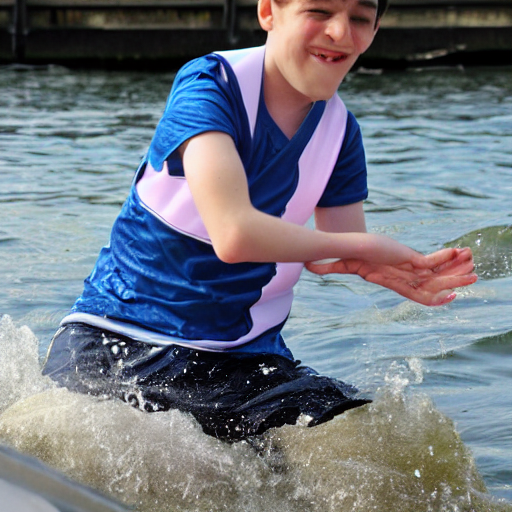}} &
 \raisebox{-18pt}{\includegraphics[height=0.12\textwidth, width = 0.12\textwidth]{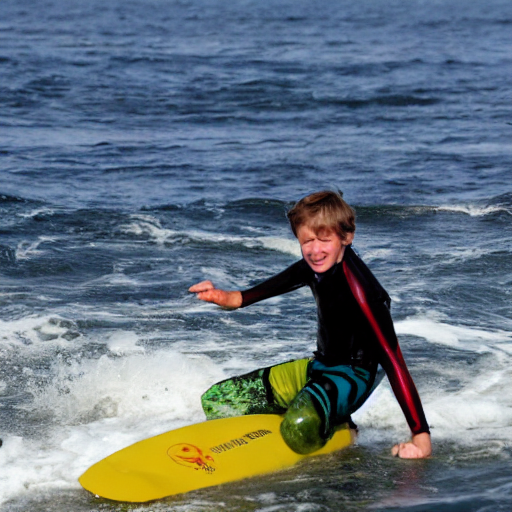}} &
 \raisebox{-18pt}{\includegraphics[height=0.12\textwidth, width = 0.12\textwidth]{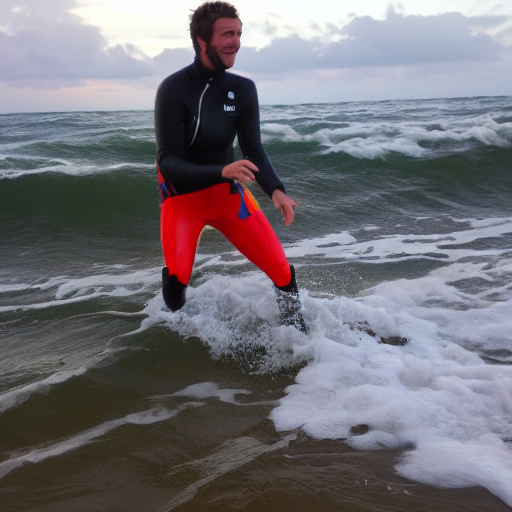}}&
 \raisebox{-18pt}{\includegraphics[height=0.12\textwidth, width = 0.12\textwidth]{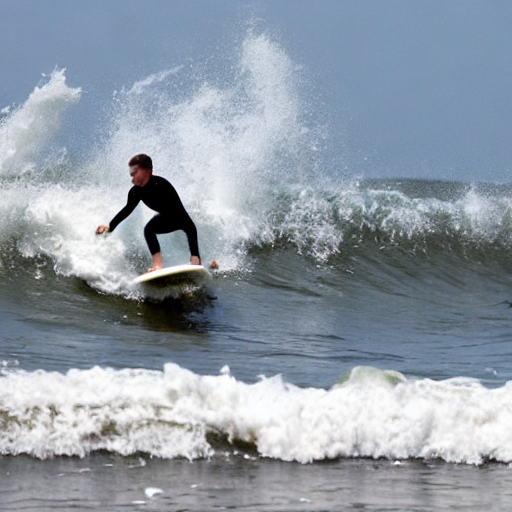}}\\
& \multicolumn{3}{m{0.36\textwidth+3\tabcolsep+\arrayrulewidth\relax}}{rts disability featured ,@ nyc autism / kidpatrick sinai kf enjoys grasp hugging otw waves } & \multicolumn{3}{m{0.36\textwidth+3\tabcolsep+\arrayrulewidth\relax}}{toby enjoying rough navigate surfing surf wave implications}\\
 \midrule
  \raisebox{-18pt}{\includegraphics[height=0.12\textwidth, width = 0.12\textwidth]{sec/figures/image_similarity/38.jpg}}&
 \raisebox{-18pt}{\includegraphics[height=0.12\textwidth, width = 0.12\textwidth]{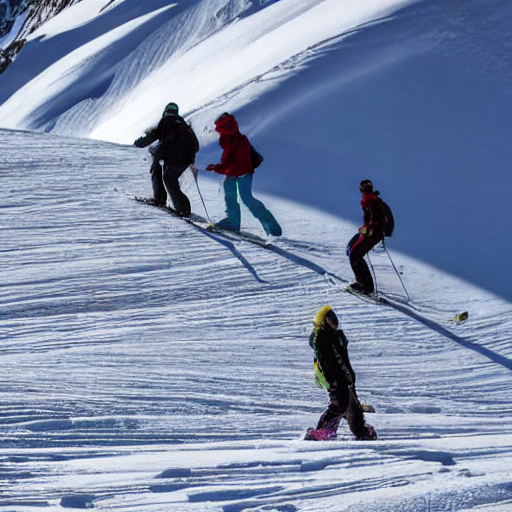}} &
 \raisebox{-18pt}{\includegraphics[height=0.12\textwidth, width = 0.12\textwidth]{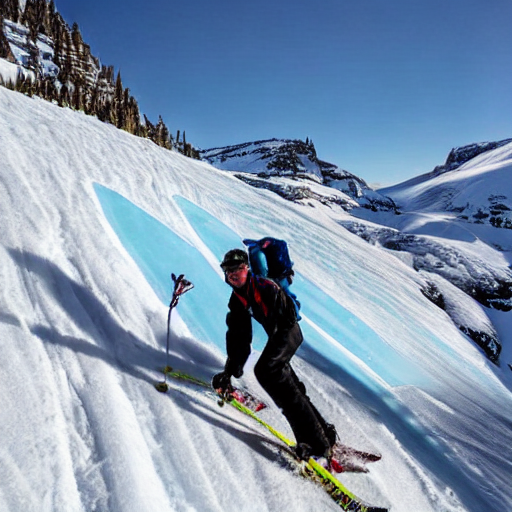}} &
 \raisebox{-18pt}{\includegraphics[height=0.12\textwidth, width = 0.12\textwidth]{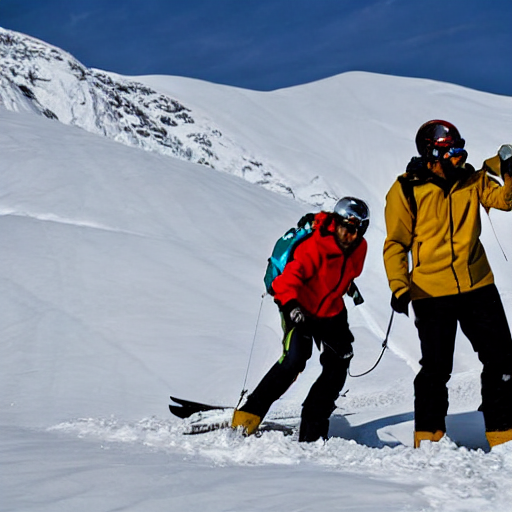}} &
 \raisebox{-18pt}{\includegraphics[height=0.12\textwidth, width = 0.12\textwidth]{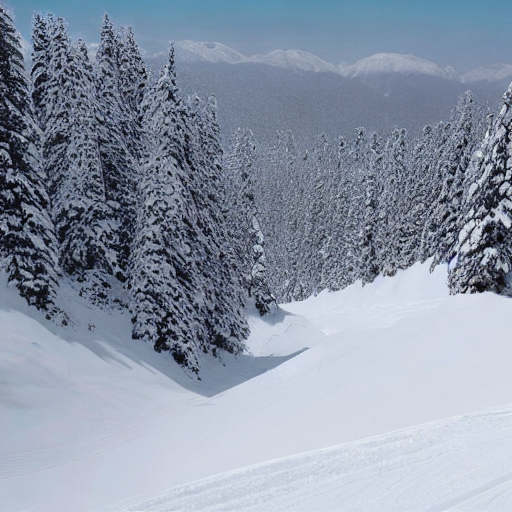}} &
 \raisebox{-18pt}{\includegraphics[height=0.12\textwidth, width = 0.12\textwidth]{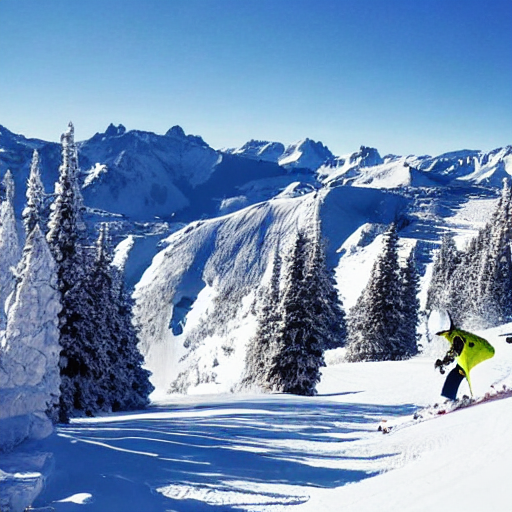}}&
 \raisebox{-18pt}{\includegraphics[height=0.12\textwidth, width = 0.12\textwidth]{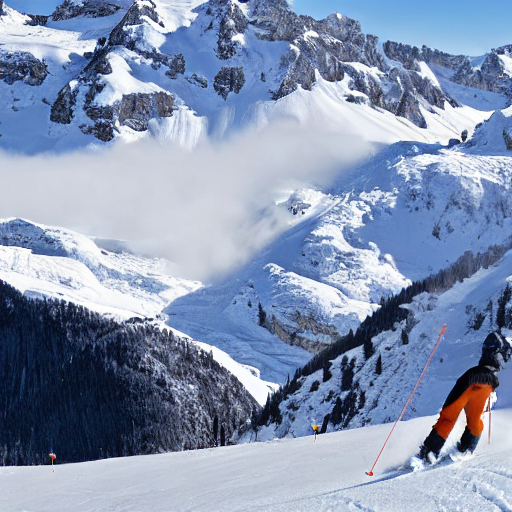}}\\
 & \multicolumn{3}{m{0.36\textwidth+3\tabcolsep+\arrayrulewidth\relax}}{bbloventureaktail vows skipped glacier corrie you <<< pow tour guided vinci agronhikes} & \multicolumn{3}{m{0.36\textwidth+3\tabcolsep+\arrayrulewidth\relax}}{alps tux beautiful skiing else S after resort}\\
 \midrule
  \raisebox{-18pt}{\includegraphics[height=0.12\textwidth, width = 0.12\textwidth]{sec/figures/image_similarity/39.jpg}}&
 \raisebox{-18pt}{\includegraphics[height=0.12\textwidth, width = 0.12\textwidth]{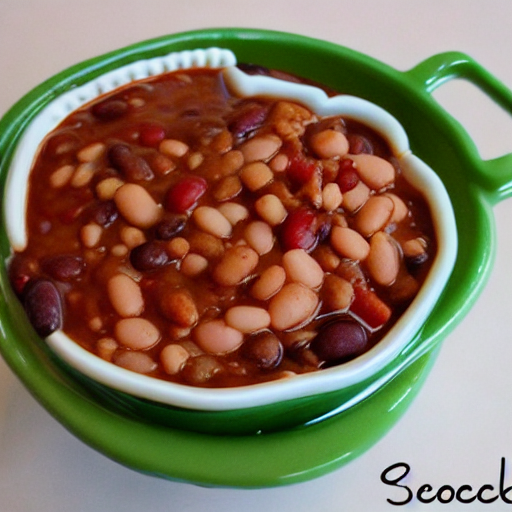}} &
 \raisebox{-18pt}{\includegraphics[height=0.12\textwidth, width = 0.12\textwidth]{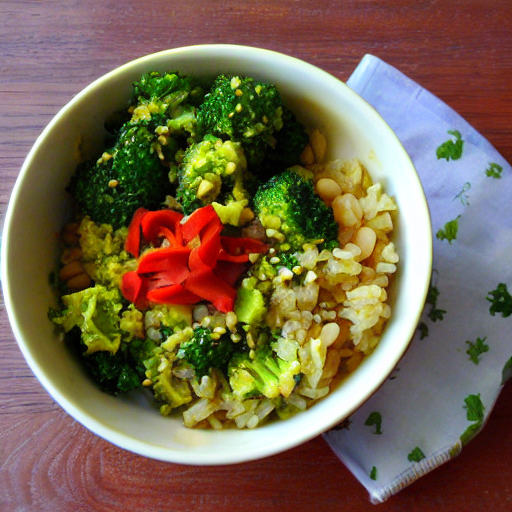}} &
 \raisebox{-18pt}{\includegraphics[height=0.12\textwidth, width = 0.12\textwidth]{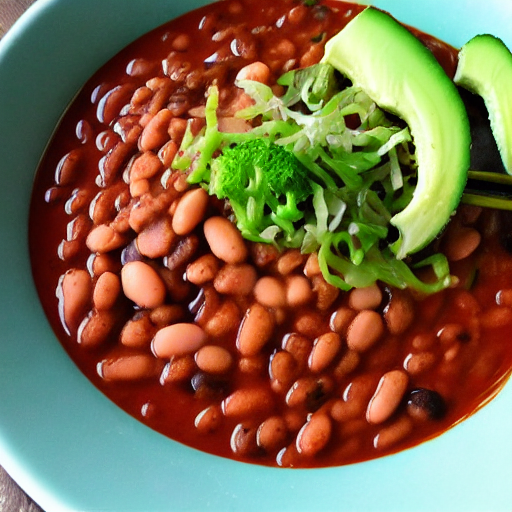}} &
 \raisebox{-18pt}{\includegraphics[height=0.12\textwidth, width = 0.12\textwidth]{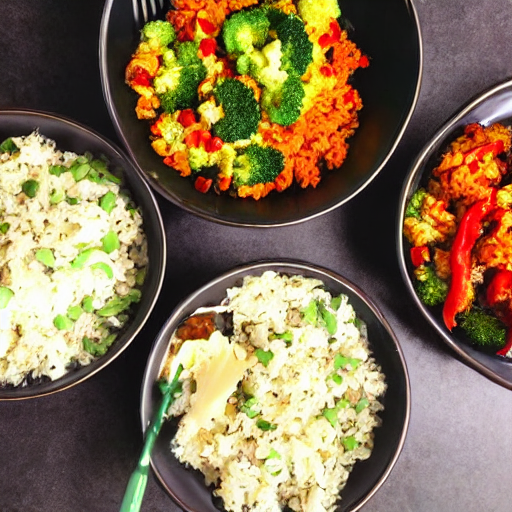}} &
 \raisebox{-18pt}{\includegraphics[height=0.12\textwidth, width = 0.12\textwidth]{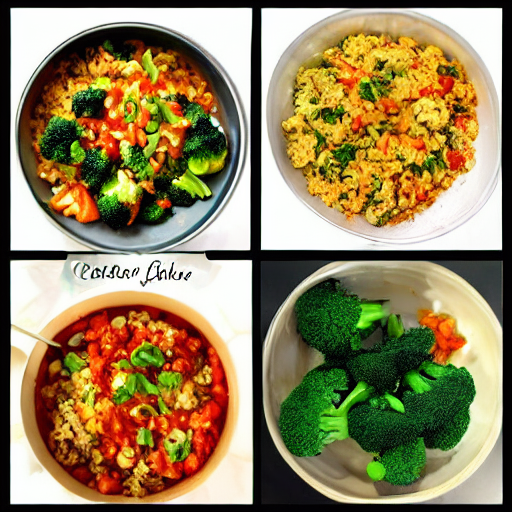}}&
 \raisebox{-18pt}{\includegraphics[height=0.12\textwidth, width = 0.12\textwidth]{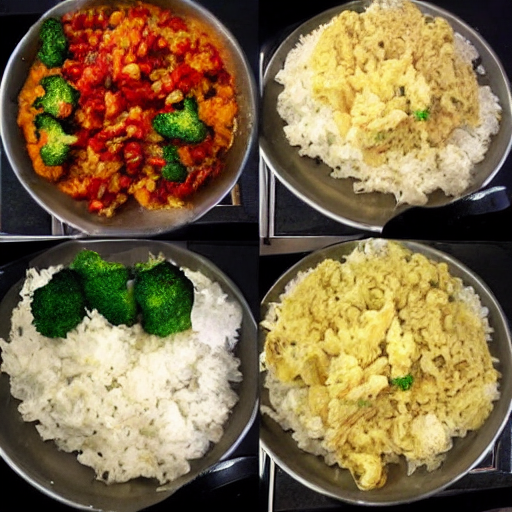}}\\
 & \multicolumn{3}{m{0.36\textwidth+3\tabcolsep+\arrayrulewidth\relax}}{\img{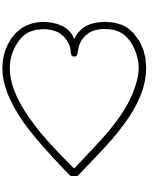}blogged clegg pulls paleo jaredbroccoli bres '? " beans protein omo chili \img{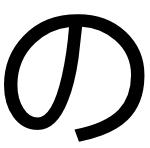}} & \multicolumn{3}{m{0.36\textwidth+3\tabcolsep+\arrayrulewidth\relax}}{homemade cauliflower chilli meals broccoli ashi rice}\\
\bottomrule
\end{tabularx}
\caption{{\bf Qualitative Comparison}. Target image, inverted prompts (from PEZ \cite{WenHPME23} and PH2P), and corresponding generated images.}
\label{tab:qualitative_PEZ_PH2P}
\vspace{-5 mm}
\end{table*}
\begin{table*}
 \centering
 \scriptsize
\begin{tabularx}{\textwidth}{@{}*{9}{m{0.085\textwidth}}@{}}
\toprule
Target Image & \multicolumn{8}{l}{Attented regions in the target image for PH2P prompt inversion} \\
\midrule
  \raisebox{-8pt} {\includegraphics[height=0.10\textwidth, width = 0.10\textwidth]{sec/figures/attention/201_000000020247.jpg}} &
 \raisebox{-18pt} {\includegraphics[width = 0.10\textwidth,keepaspectratio]{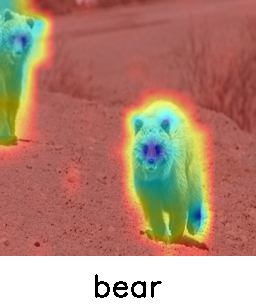}} &
\raisebox{-18pt}{\includegraphics[width = 0.10\textwidth,keepaspectratio]{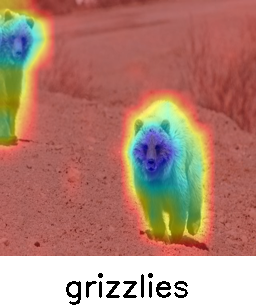}} &
\raisebox{-18pt}{\includegraphics[width = 0.10\textwidth,keepaspectratio]{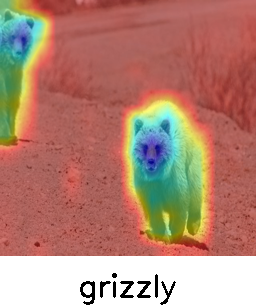}} &
\raisebox{-18pt}{\includegraphics[width = 0.10\textwidth,keepaspectratio]{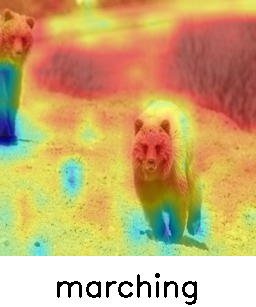}} &
\raisebox{-18pt}{\includegraphics[width = 0.10\textwidth,keepaspectratio]{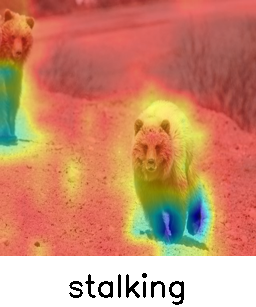}}&
\raisebox{-18pt}{\includegraphics[width = 0.10\textwidth,keepaspectratio]{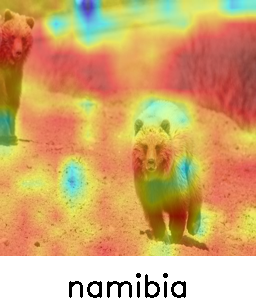}}
&
\raisebox{-18pt}{\includegraphics[width = 0.10\textwidth,keepaspectratio]{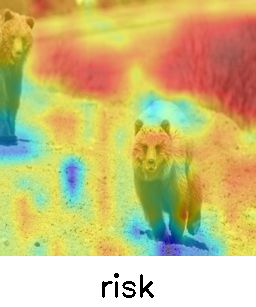}}
&
\raisebox{-18pt}{\includegraphics[width = 0.10\textwidth,keepaspectratio]{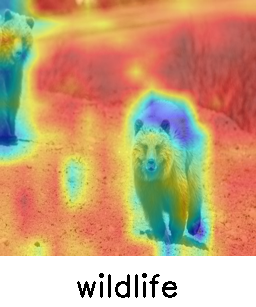}}
\\
\midrule
  \raisebox{-8pt} {\includegraphics[height=0.10\textwidth, width = 0.10\textwidth]{sec/figures/attention/39_000000003501_2.jpg}} &
 \raisebox{-18pt} {\includegraphics[ width = 0.10\textwidth,keepaspectratio]{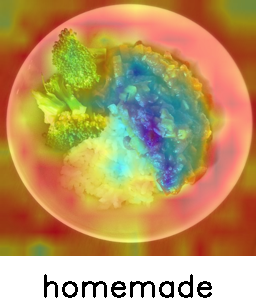}} &
\raisebox{-18pt}{\includegraphics[width = 0.10\textwidth,keepaspectratio]{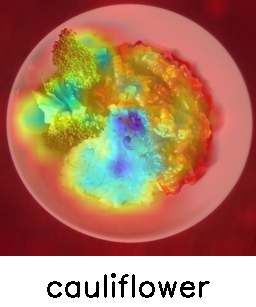}} &
\raisebox{-18pt}{\includegraphics[width = 0.10\textwidth,keepaspectratio]{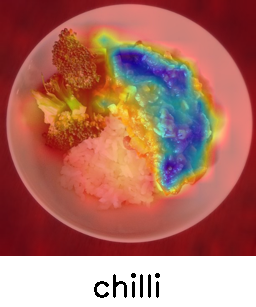}} &
\raisebox{-18pt}{\includegraphics[width = 0.10\textwidth,keepaspectratio]{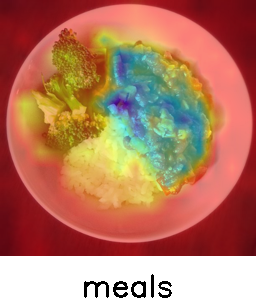}} &
\raisebox{-18pt}{\includegraphics[width = 0.10\textwidth,keepaspectratio]{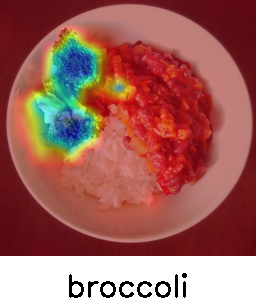}}&
\raisebox{-18pt}{\includegraphics[width = 0.10\textwidth,keepaspectratio]{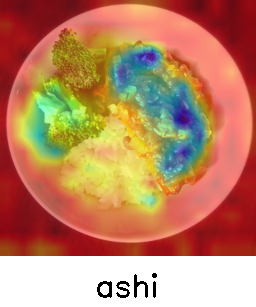}}
&
\raisebox{-18pt}{\includegraphics[width = 0.10\textwidth,keepaspectratio]{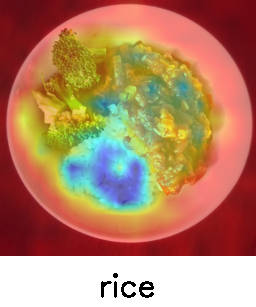}}
&\\
\bottomrule
\end{tabularx}
\vspace{-2 mm}
\captionof{figure}{{\bf Application of Unsupervised Segmentation.} Illustration of the tokens and corresponding regions (that can be used for unsupervised segmentation; see \cite{abs-2303-11681}) obtained for the target images. Note the accuracy of both prompts and corresponding attention.} 
\label{tab:attention_maps}
\vspace{-3 mm}
\end{table*}

We empirically demonstrate our prompt inversion procedure on the COCO \cite{LinMBHPRDZ14} and the SUN \cite{XiaoHEOT10} datasets. 
We validate our results on 500 images from the validation sets across 5 seeds and use Stable Diffusion v1.5 as the pre-trained diffusion model.

\paragraph{Evaluation metrics.} 
We consider the following \ky{three} aspects when evaluating the quality of inverted prompts: 
\begin{itemize}
    \item Accuracy of the inverted prompts: we measure the CLIP \cite{RadfordKHRGASAM21} and the LPIPS \cite{ZhangIESW18} image similarity scores to measure the semantic similarity between the target images and the generated images. 
    \item Diversity of the generated images: we measure the diversity of the generated images using LPIPS between different generated samples.
    \item Interpretability of the inverted prompts: we use  the BertScore~\cite{ZhangKWWA20}, which correlates well with human judgment. We measure the semantic equivalence between the embeddings of two sentences by taking into account the context of the tokens. 
\end{itemize}
%
\begin{figure}
    \centering
    \begin{subfigure}{0.48\textwidth}
    \includegraphics[width=\textwidth]{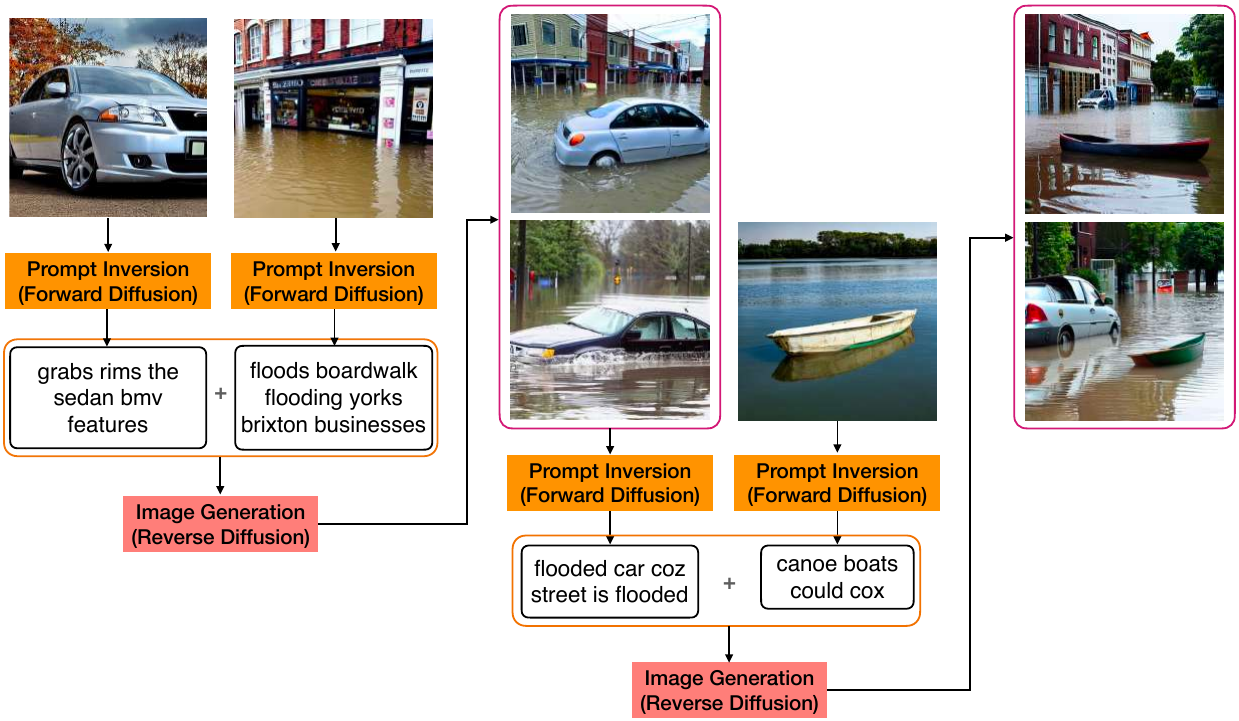}
    \end{subfigure} \\
    \begin{subfigure}{0.48\textwidth}
    \includegraphics[width=\textwidth]{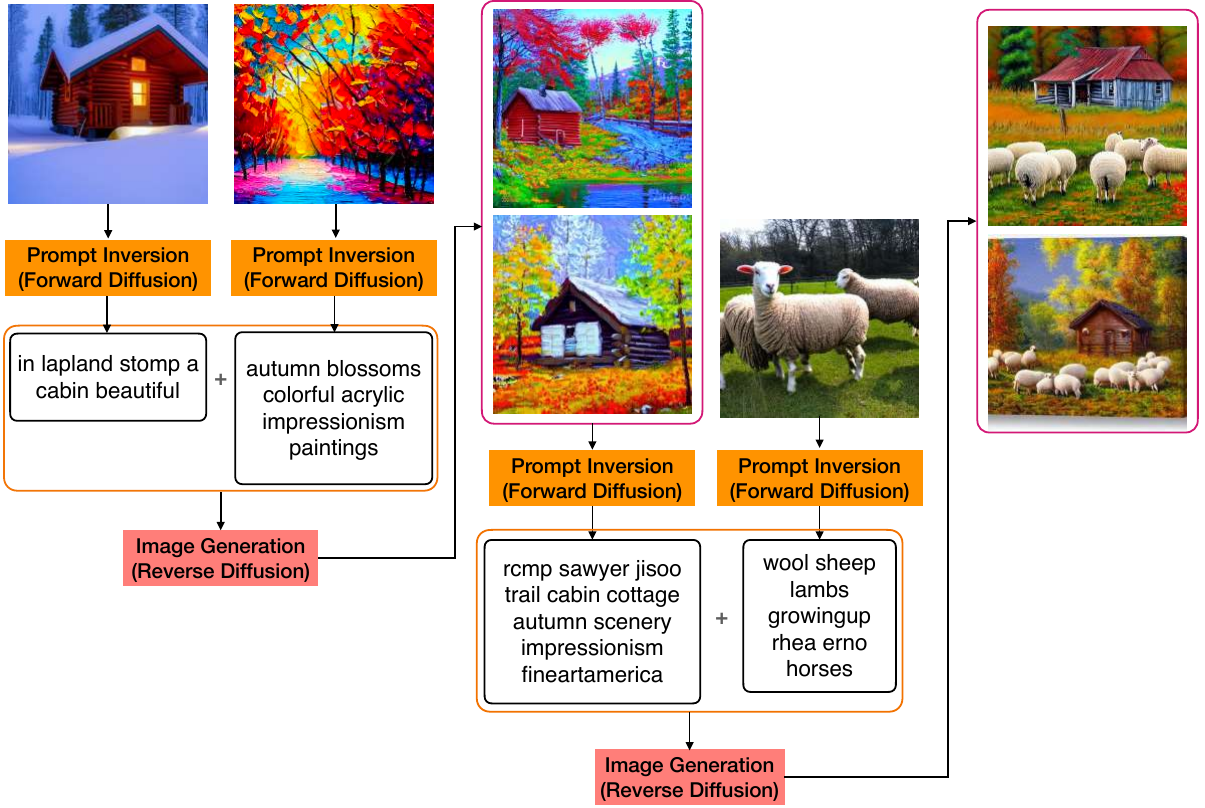}
    \end{subfigure}
    \caption{{\bf Application of Evolutionary Multi-concept Generation with Proposed PH2P.} Images generated with the composed prompts are inverted with PH2P to create prompts which are further combined with prompts recovered from the new target image.}
    \label{fig:evolutionary_generation}
    \vspace{-0.35cm}
\end{figure}

\subsection{Evaluation of the Inverted Prompts}

\vspace{-\customparskip}
\paragraph{Image generation with inverted prompts.}
We assess the effectiveness of the prompts generated using our PH2P inversion for image generation. 
Given a target image, we first perform inversion to retrieve the prompt and evaluate the relevance of the images generated with the recovered prompt using the diffusion model. 
We compare our approach to the prompts generated using (concurrent/unpublished) PEZ \cite{WenHPME23} 
as a baseline. 
As shown in \cref{table:coco_sun_quant}, the images generated with our PH2P prompts outperform the PEZ prompts in terms of CLIP similarity by an absolute value of $5$ percentage points and $7$ percentage points on the COCO and SUN datasets, respectively. 
Furthermore, our PH2P prompts display better performance in terms of the LPIPS similarity. 
This demonstrates the accuracy and relevance of the PH2P prompts for generating images with visual concepts of intent.

When comparing the diversity of the images generated with our PH2P versus PEZ \cite{WenHPME23}, our PH2P approach yields higher diversity while maintaining the image semantics.
Our results highlight that our approach recovers the prompts in the space of the model's vocabulary; these prompts can be used to sample diverse yet relevant images with visual concepts aligned with those of the target image.

Extensive qualitative results in \cref{tab:qualitative_PEZ_PH2P} show that our approach generates prompts representative of the visual content in the target images. 
Unlike PEZ \cite{WenHPME23}, where the generated images, at times, cannot capture the intended target concepts (\eg, row 2), our approach consistently generates images with accurate semantics. 
\begin{figure*}
\centering
    \includegraphics[width=0.65\textwidth]{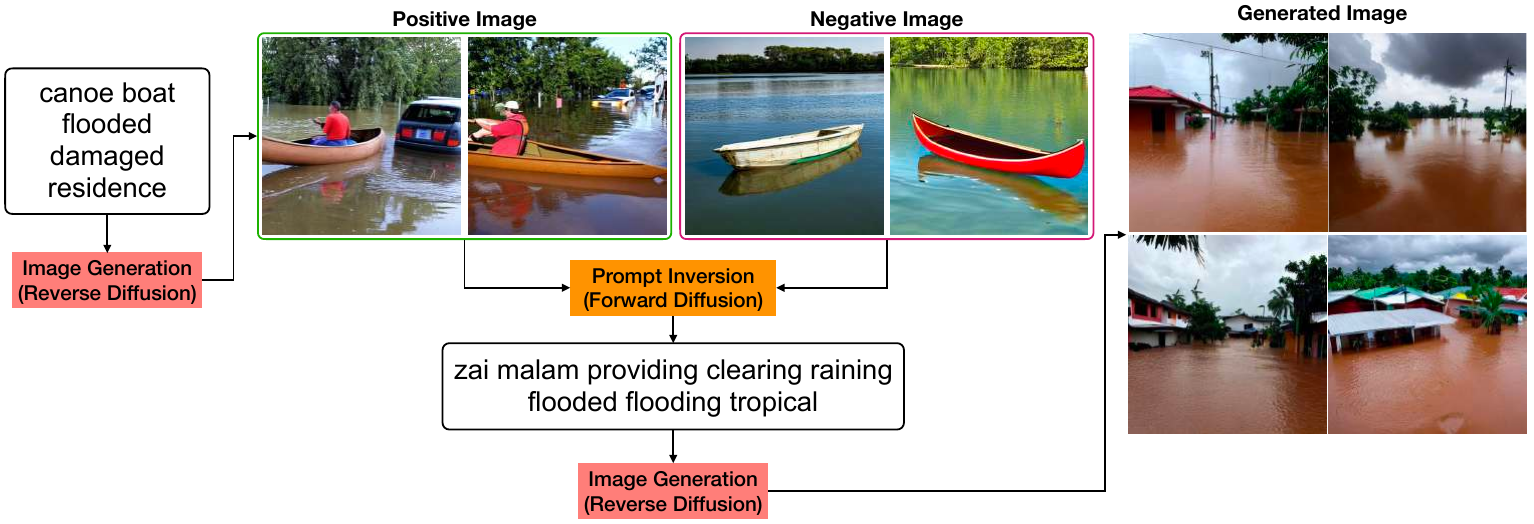}
\vspace{-2 mm}
\caption{{\bf Application of Concept Removal with Negative Target Images.} PH2P yields a prompt that removes the visual concept given in the negative image from the positive target image. The PH2P prompts can be used to generate diverse images with removed concepts.}
\label{fig:negative_prompt}
\vspace{-5 mm}
\end{figure*}
\begin{table}
\scriptsize
\centering
  \begin{tabularx}{\linewidth}{@{}Xccc@{}}
  \toprule
   Method () & Precision($\uparrow$) & Recall ($\uparrow$) & F1 ($\uparrow$)\\
    \midrule
    PEZ \cite{WenHPME23} & 0.772 & 0.835 &0.802 \\
    PH2P (Ours) & {\bf 0.803} & {\bf 0.838} & {\bf 0.820} \\
    \bottomrule
  \end{tabularx}
  \vspace{-2mm}
  \caption{{\bf Quantitative Evaluation of Prompt Quality.} Evaluation of quality of prompts generated on the COCO dataset as captured bu the BertScore \cite{ZhangKWWA20}. We compare the similarity between the inverted prompts and the ground-truth captions.}
  \label{tab:quantitative_bertscore}
  \vspace{-0.2cm}
\end{table}

\paragraph{Quality of the prompts.}
We compare the contextual similarity between the prompts inverted for the COCO dataset and the ground-truth annotations (captions) of the corresponding images, with the BertScore \cite{ZhangKWWA20}.
Results in \cref{tab:quantitative_bertscore} show that our approach outperforms the prompts generated by PEZ \cite{WenHPME23}  in terms of BertScore, especially with respect to the precision evaluation validating that our PH2P prompts have greater semantic similarity to the human captions. 
Our PH2P provides relatively precise prompts compared to those from PEZ and has better contextual similarity to the ground-truth captions.
We observe from \cref{tab:qualitative_PEZ_PH2P} that the prompts from our PH2P approach are more crisp and clear compared to those from PEZ \cite{WenHPME23}. 
The prompts generated with PEZ tend to have a high frequency of special or uninterpretable characters; see~rows 2, 4, 6 in \cref{tab:qualitative_PEZ_PH2P}.

\begin{table}
\scriptsize
\centering
  \begin{tabularx}{\linewidth}{@{}Xccc@{}}
  \toprule
   Method & CLIP Similarity($\uparrow$) & LPIPS Similarity ($\downarrow$) & LPIPS Diversity ($\uparrow$)\\
    \midrule
      PH2P & {\bf 0.77} & {\bf 0.462} & {\bf 0.435}\\
      LDM+adam & 0.65 & 0.501 &0.400\\
      LDM+all $t$ & 0.72 & 0.479 &0.423\\
    \bottomrule
  \end{tabularx}
  \vspace{-2mm}
  \caption{{\bf Ablation.} Ablations study the significance of optimization choices for inversion in PH2P.}
  \label{table:ablations}
 \vspace{-5 mm}
\end{table}

\paragraph{Ablations.} 
To justify the optimization choices for PH2P prompt inversion in diffusion models, we show in \cref{table:ablations} the performance of prompt inversion with Adam optimizer and with optimization for all timesteps (as opposed to the selected range). 
When using Adam optimizer, the approach yields prompts that generate images with relatively low CLIP and LPIPS similarity. 
The Adam optimizer with a fixed learning rate often does not converge to a good solution.
Similarly, when optimizing with all the timesteps even though, the CLIP similarity of the generated images with recovered prompts is comparable to that of PEZ, due to high variance in the trajectory the results lag behind our proposed PH2P procedure.

\subsection{Applications of Prompt Inversion}
In this section, we show the benefits of prompt inversion in evolutionary multi-concept synthesis and in concept removal via negative image prompting.

\paragraph{Evolutionary multi-concept image synthesis.}
Consider an application where the goal of the user is to synthesize a mentally constructed image (\eg, {\em an image of a flooded city}). A convenient way to do so, would be to select some sample images (\eg, obtained from Google image search) that contain concepts he/she wants to see (\eg, {\em car}, {\em flooded street}). These images can be used to prompt the text-to-image generative model to generate desired illustration. Once the target images are generated, a user may realize that inserting a boat may give a more striking impact and would want to incorporate that concept. Such creative process is fundamentally enabled by our prompt inversion. Consider the two examples in \cref{fig:evolutionary_generation}. 
In the first step the concepts of two (or more) images can be composed by performing a prompt inversion of each image (set) and subsequently concatenating the prompts of the two (sets) of images. 
In step two, the newly composed images generated with the concatenated prompt can further be inverted to get a precise prompt that takes into account the text prior within the diffusion models. 
The inverted prompt in step 2 can further be combined with the inverted prompt of a new image depicting an additional concept. 
This allows convenient multi-concept composition (also enabling user prompt editing).

\paragraph{Negative image prompting.}
Our PH2P algorithm allows for the removal of concepts from images through negative image prompting; see~\cref{fig:negative_prompt}.
Given an input prompt, a set of positive images is generated representing the semantics of the input prompt. 
To remove a visual concept from the positive images, a negative target image with the negative visual concept is presented. 
To do this, we adapt our PH2P procedure and include negative gradients for the negative target image, generating prompts that drive towards the concepts in positive images and at the same away from the concepts in the negative target image; see supplementary appendix for more details.

\paragraph{Unsupervised Segmentation.}
Another application of prompt inversion is to generate results for unsupervised semantic segmentation (see \cite{abs-2303-11681}) where the cross-attention maps are used to generate segmentation masks. 
With our approach, one can generate prompts 
for the representative concepts directly from the diffusion model without any external information.
For a given target image, we first perform PH2P inversion to generate such
prompts. 
We visualize in \cref{tab:attention_maps} the cross-attention maps between the target image and the generated prompts \cite{CheferAVWC23}.
Clearly, the tokens reflect the concepts relevant to the image,
accurately attend to corresponding image, and can generate segmentations.
\section{Conclusion}

In this paper, we designed a simple but effective procedure for prompt inversion in text-to-image diffusion models. 
The prompts generated with our approach are readable, crisp, and importantly, representative of the content of the target image. 
We demonstrate the effectiveness of our inverted prompts for diverse image generation, evolutionary concept generation, and even concept removal. 
We believe that our inversion procedure would make prompting diffusion models 
much easier 
by allowing users to 
construct (and edit) effective textual descriptions of concepts they desire.

\section*{Acknowledgments}
\small
This work was funded, in part, by the Vector Institute for AI, Canada CIFAR AI Chairs, Natural Sciences and Engineering Research Council of Canada (NSERC) Discovery Grant, NSERC Collaborative Research and Development Grant. Resources used in preparing this research were provided, in part, by the Province of Ontario, the Government of Canada through CIFAR, the Digital Research Alliance of Canada \url{alliance.can.ca}, \href{https://vectorinstitute.ai/\#partners}{companies} sponsoring the Vector Institute, and Advanced Research Computing at the University of British Columbia. Additional hardware support was provided by John R. Evans Leaders Fund CFI grant and Compute Canada under the Resource Allocation Competition award.

\small
\bibliographystyle{ieeenat_fullname}
\bibliography{main}


\clearpage
\appendix
\appendixpage
\addappheadtotoc

In the following, we outline the algorithm for the negative image prompting for prompt inversion and also provide additional qualitative results to show that the prompts generated with our PH2P approach are precise, readable, and can be used for various downstream applications.

\section{Negative Prompting}
In \Cref{algo:neg_samp:supp}, we show the adaptation of our PH2P \cref{algo:samp} to get prompts that contain the concepts from a target image and at the same time do not contain concepts specified in the negative image. 
The images thus generated with the optimized prompt contain concepts in $\mathbf{I}$ excluding or removing the concepts present in $\mathbf{I}^{\text{neg}}$. 
We find the algorithm to perform well when removing concepts in composed images (\cref{fig:negative_prompt,fig:negative_prompting:supp}).

\begin{algorithm}[h]
\scriptsize
\SetNoFillComment
Input: Diffusion model parameters: $\theta$, Target image: $\mathbf{x}=E(\mathbf{I})$, Negative image: $\mathbf{x}^{\text{neg}}=E(\mathbf{I}^{\text{neg}})$, Initial prompt: $\mathbf{S}$, Prompt embedding: $\mathbf{\hat{e}}$,
Timesteps: $[t_a, T]$; Learning rate: $\lambda$, Optimization steps: $N$\\
 \For{$i\gets 1 $ \KwTo $N$}{
 \tcc{Projection on feasible set}
        $\tilde{\mathbf{e}} =\text{Proj}_\mathbf{E}(\hat{\mathbf{e}}) $ \\
 \tcc{Select diffusion timestep}
        $t = \text{random}([t_a,T])$\\
  \tcc{Apply L-BFGS}
      $g=\text{LBFGS}_{\tilde{\mathbf{e}}}\left(\mathcal{L}_{LDM}(\mathbf{x}_t,\theta,t,f(\tilde{\mathbf{e}}))-\mathcal{L}_{LDM}(\mathbf{x}^{\text{neg}}_t,\theta,t,f(\tilde{\mathbf{e}}))\right)$
        \\
        $\hat{\mathbf{e}} = \hat{\mathbf{e}}-\lambda g$
        }
\tcc{Delayed projection}
return $\text{Proj}_\mathbf{E}(\hat{\mathbf{e}})$   

 \caption{PH2P Prompt Inversion with Negative Image Prompting}
 \label{algo:neg_samp:supp}
\end{algorithm}

\section{Additional Qualitative Results}

In \cref{tab:qualitative_PEZ_PH2P:supp} we provide additional qualitative results comparing the quality of the prompts from our PH2P approach with the PEZ approach based on CLIP similarity. 
Here, we observe consistently better prompts. 
Images generated with our inverted prompts reflect the concepts of target image.

Additionally, we show in \cref{tab:ablations:supp} the prompts generated with the standard Adam optimization (LDM+Adam) and also consider the setting with all the timesteps for inversion (LDM+ all $t$). 
We observe that the prompts generated with our PH2P procedure are consistent with the concepts in the target image. 
The prompts with LDM+Adam are short but do not always contain the target concepts.
This shows the poor convergence of the standard SGD methods for hard prompt inversion. 
The prompts obtained with LDM + all steps are longer than our PH2P procedure and tend to include words that do not provide any information on the concepts in an image.
Our PH2P prompt inversion, on the other hand, provides prompts that are readable, precise, and consistent with the content of the image.

Similarly, in \cref{tab:attention_maps:supp} we show the regions in the target image attended to by the optimized prompts which can be leveraged for downstream tasks such as unsupervised semantic segmentation as proposed in Wu \etal~\cite{abs-2303-11681}.

\begin{figure*}
    \centering
    \begin{subfigure}{0.8\textwidth}
    \includegraphics[width=\textwidth]{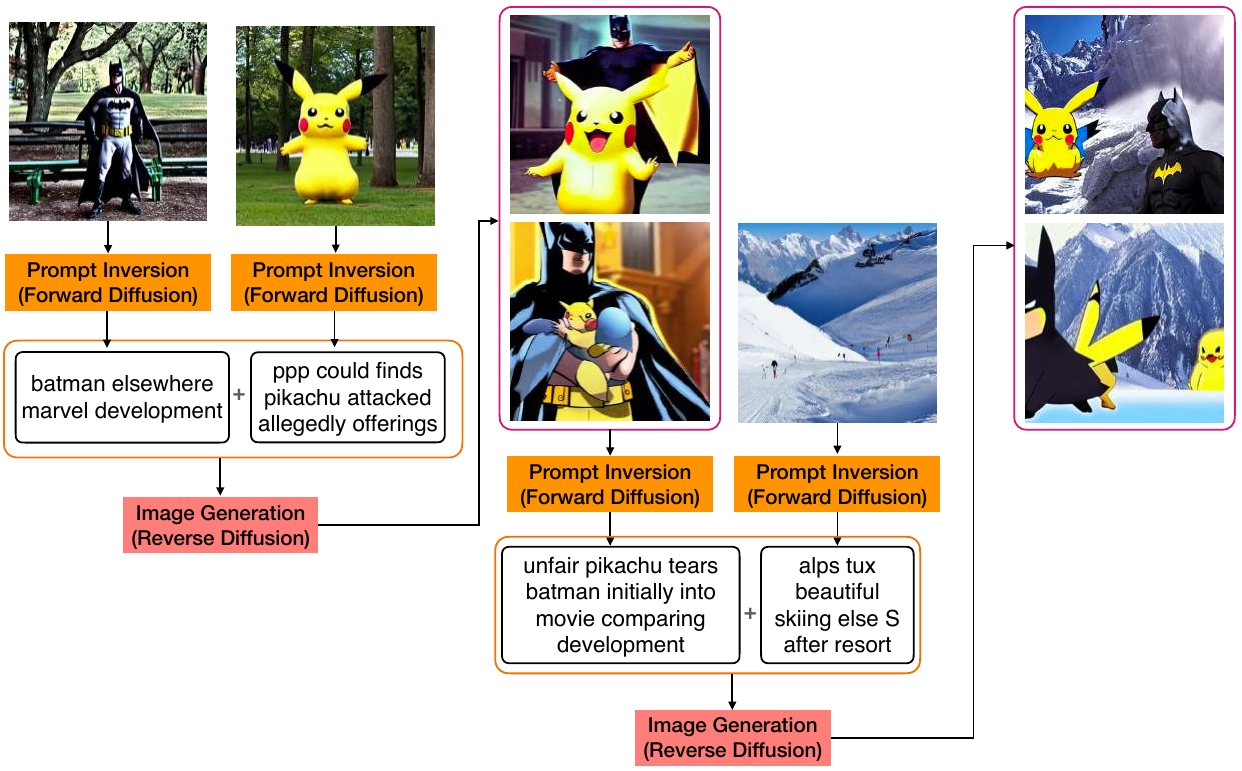}
    \end{subfigure} \\
  
    \begin{subfigure}{0.8\textwidth}
    \includegraphics[width=\textwidth]{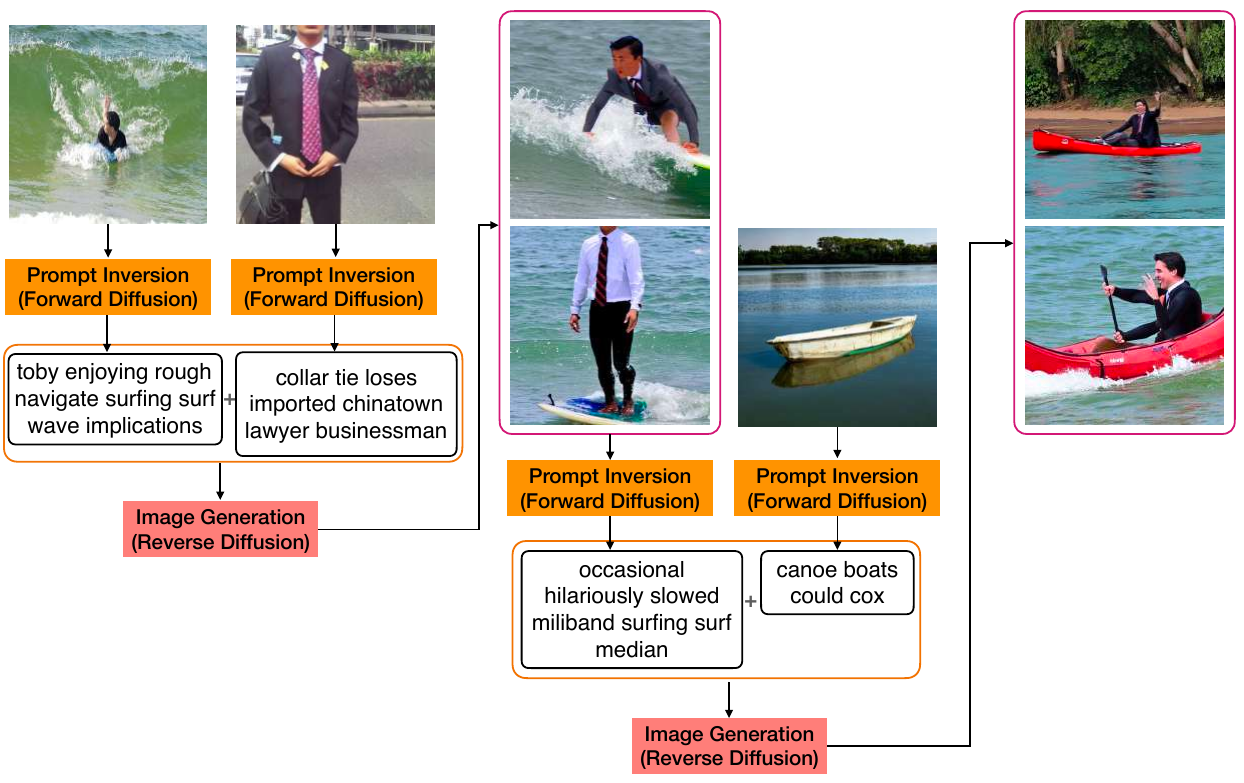}
    \end{subfigure}
    \caption{{\bf Application of Evolutionary Multi-concept Generation with Proposed PH2P.} Additional results on images generated with the composed prompts are inverted with PH2P to create prompts which are further combined with prompts recovered from the new target image.}
    \label{fig:evolutionary_generation:supp}
    \vspace{-0.35cm}
\end{figure*}

\begin{figure*}
    \centering
    \begin{subfigure}{0.8\textwidth}
    \includegraphics[width=\textwidth]{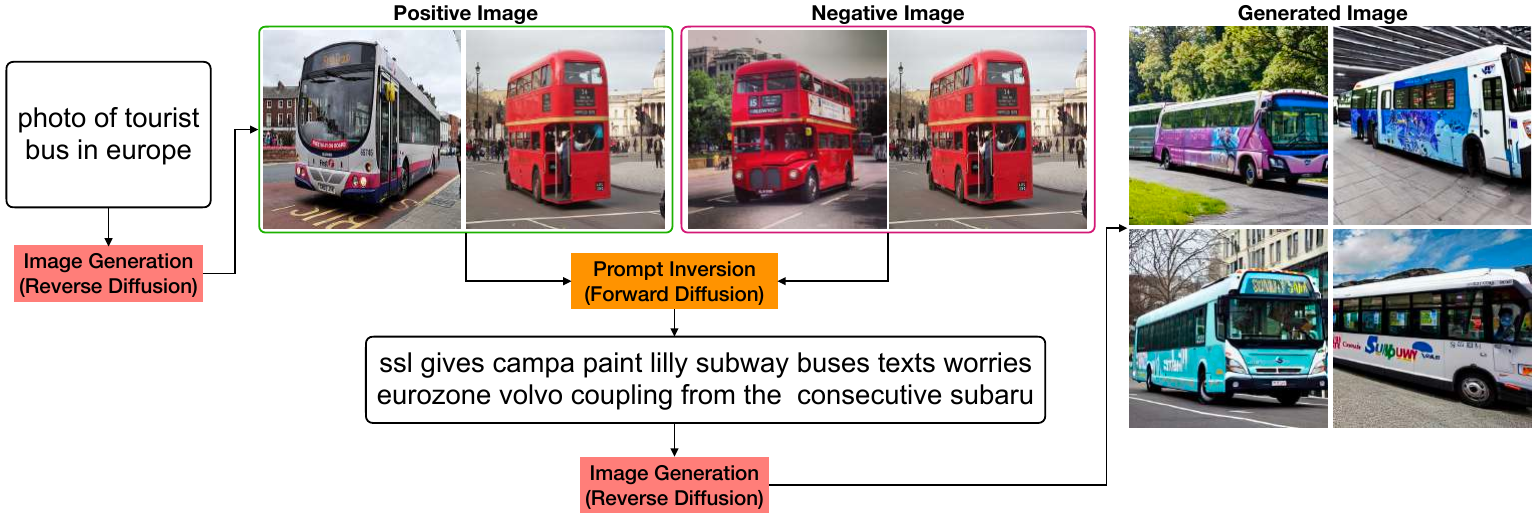}
    \end{subfigure} \\
  
    \begin{subfigure}{0.8\textwidth}
    \includegraphics[width=\textwidth]{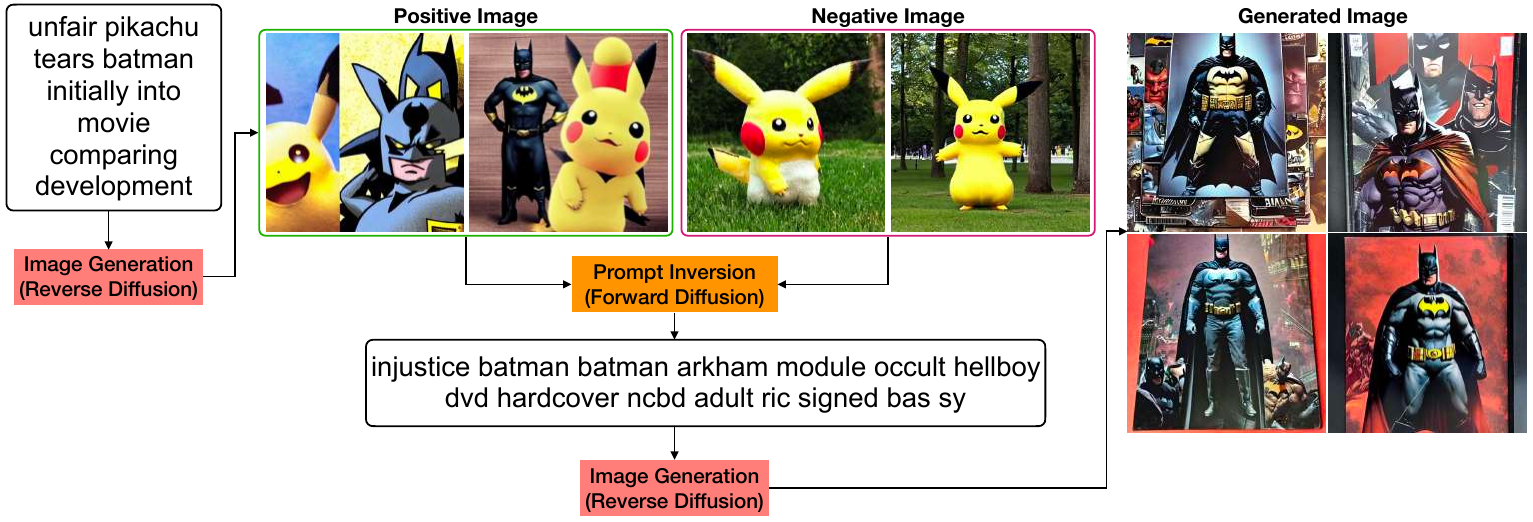}
    \end{subfigure}
    \caption{{\bf Application of Concept Removal with Negative Target Images.} Additional results showing that PH2P yields a prompt that removes the visual concept given in the negative image from the positive target image. The PH2P prompts can be used to generate diverse images with removed concepts.}
    \label{fig:negative_prompting:supp}
    \vspace{-0.35cm}
\end{figure*}

We also include additional examples for the application of the PH2P procedure to multi-concept generation \cref{fig:evolutionary_generation:supp}.
Our approach can faithfully compose images with diverse concepts.
\Cref{fig:negative_prompting:supp} shows example cases where concepts can be removed from a set of target images using our PH2P procedure.
Given the negative images of the concepts to be removed, PH2P with \cref{algo:neg_samp:supp} can be used to yield a prompt that removes concepts from target (positive) images.
These observations demonstrate that the quality of the prompts generated with the PH2P procedure is representative of the vocabulary of the LDM \cite{rombach2022high} backbone.

\begin{table*}[h]
 \centering
 \scriptsize
\begin{tabularx}{\textwidth}{@{}*{7}{m{0.12\textwidth}}@{}}
\toprule
Target Image & \multicolumn{3}{l}{PEZ \cite{WenHPME23}} & \multicolumn{3}{l}{PH2P (Ours)}\\
\cmidrule(lr){2-4} \cmidrule(lr){5-7}
  \raisebox{-18pt} {\includegraphics[height=0.12\textwidth, width = 0.12\textwidth]{sec/figures/suppmat/105.jpg}} &
 \raisebox{-18pt} {\includegraphics[height=0.12\textwidth, width = 0.12\textwidth]{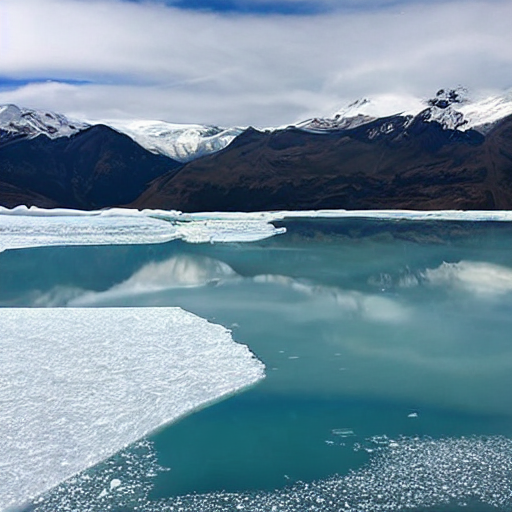}} &
\raisebox{-18pt}{\includegraphics[height=0.12\textwidth, width = 0.12\textwidth]{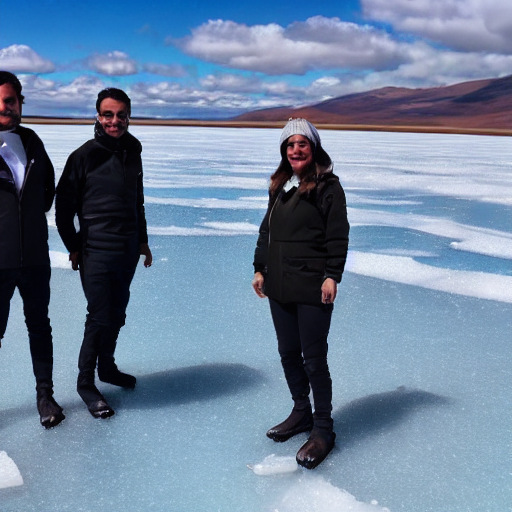}} &
\raisebox{-18pt}{\includegraphics[height=0.12\textwidth, width = 0.12\textwidth]{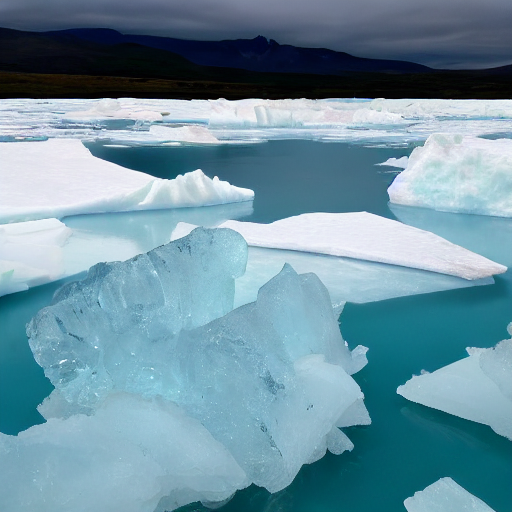}} &
\raisebox{-18pt}{\includegraphics[height=0.12\textwidth, width = 0.12\textwidth]{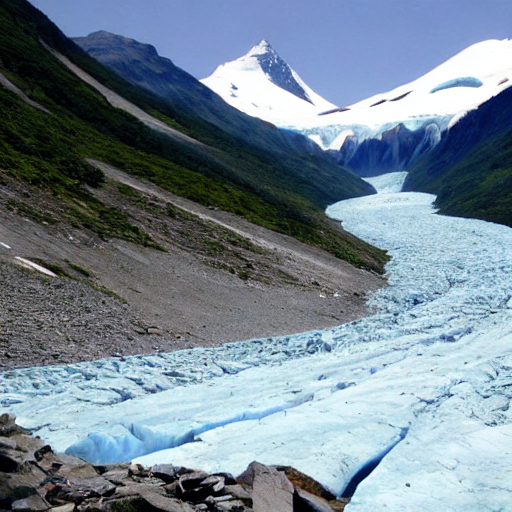}} &
\raisebox{-18pt}{\includegraphics[height=0.12\textwidth, width = 0.12\textwidth]{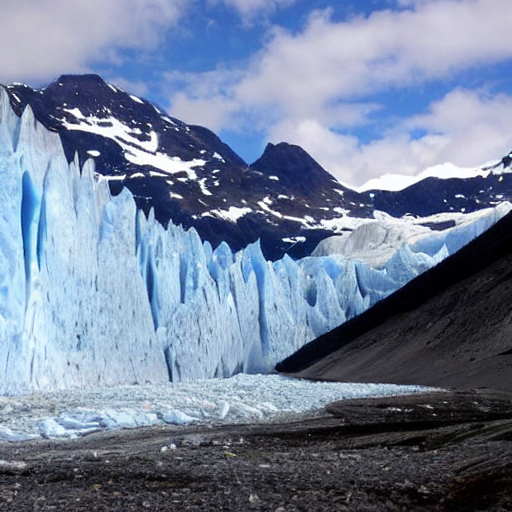}}&
\raisebox{-18pt}{\includegraphics[height=0.12\textwidth, width = 0.12\textwidth]{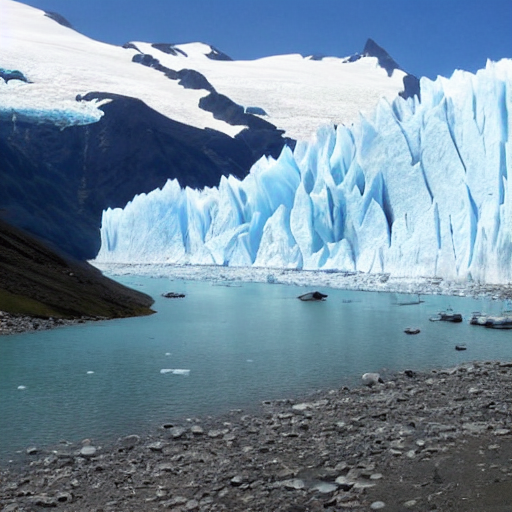}}\\
& \multicolumn{3}{m{0.36\textwidth+3\tabcolsep+\arrayrulewidth\relax}}{collaborate dreamliner shared ingrakyleargentina lake ices @$\{$ $\}$ amadiscover tos — carmenono augmente} & \multicolumn{3}{m{0.36\textwidth+3\tabcolsep+\arrayrulewidth\relax}}{glacier flows forgot milford idyllic glaciers milky forests}\\
\midrule
   \raisebox{-18pt}{\includegraphics[height=0.12\textwidth, width = 0.12\textwidth]{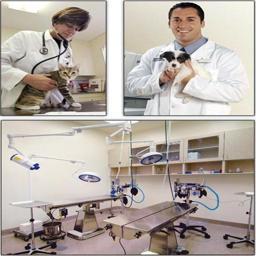}}&
 \raisebox{-18pt}{\includegraphics[height=0.12\textwidth, width = 0.12\textwidth]{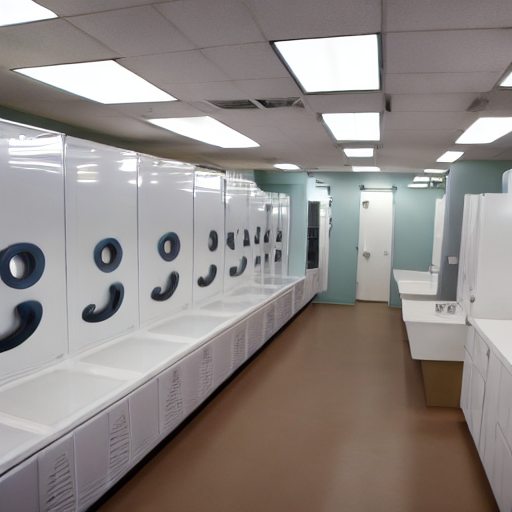}} &
 \raisebox{-18pt}{\includegraphics[height=0.12\textwidth, width = 0.12\textwidth]{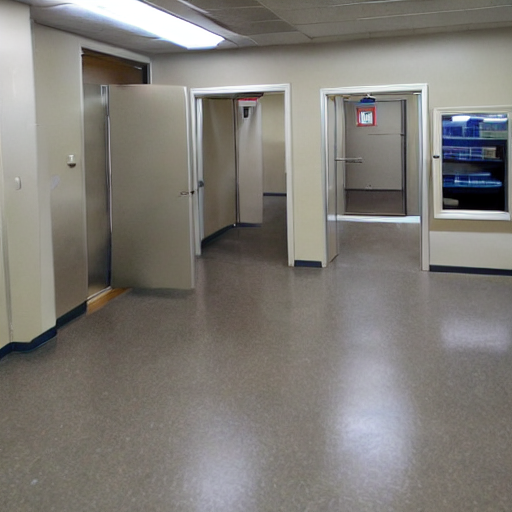}} &
 \raisebox{-18pt}{\includegraphics[height=0.12\textwidth, width = 0.12\textwidth]{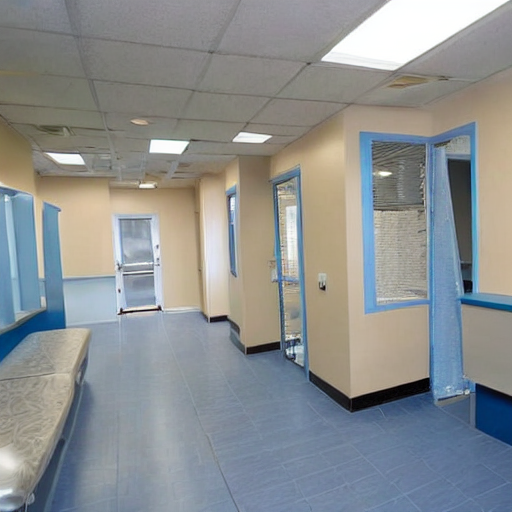}} &
 \raisebox{-18pt}{\includegraphics[height=2.0cm,width=2.0cm]{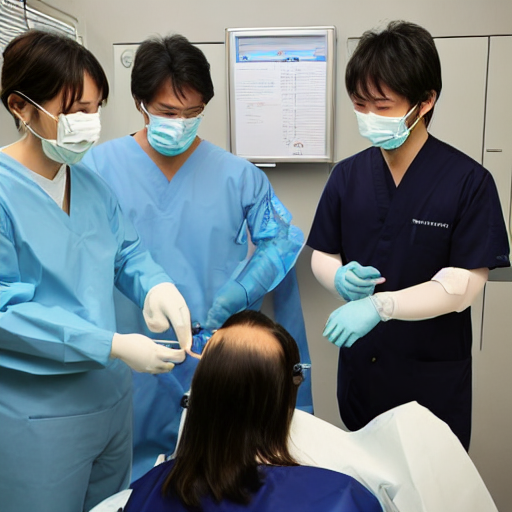}} &
 \raisebox{-18pt}{\includegraphics[height=0.12\textwidth, width = 0.12\textwidth]{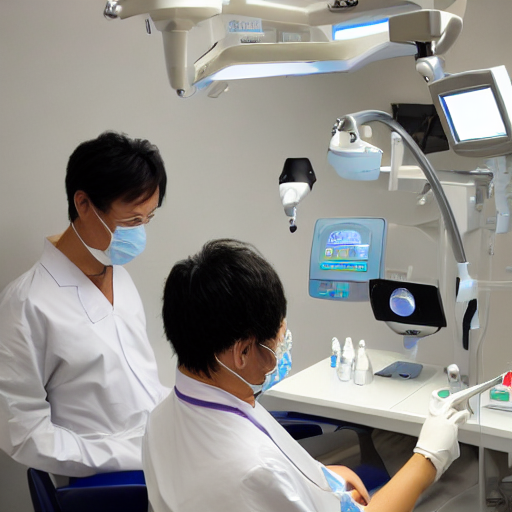}}&
 \raisebox{-18pt}{\includegraphics[height=0.12\textwidth, width = 0.12\textwidth]{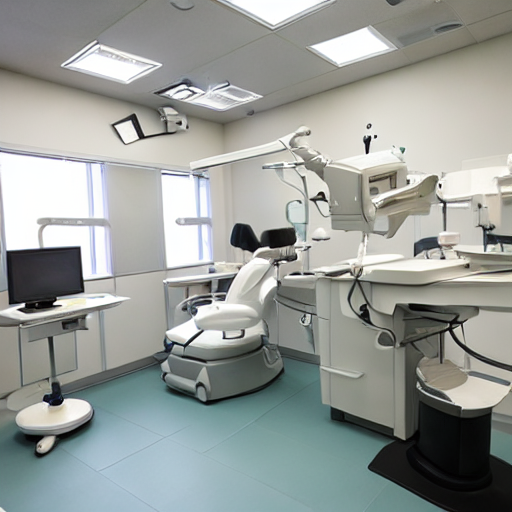}}\\
 & \multicolumn{3}{m{0.36\textwidth+3\tabcolsep+\arrayrulewidth\relax}}{permitted facilities cats dryer drexyulhomeitems veterinlished gyn extensive specializes scrubs mnt treated mnt} & \multicolumn{3}{m{0.36\textwidth+3\tabcolsep+\arrayrulewidth\relax}}{tokyo ips seals dentistry simulation surgeries}\\
 \midrule
   \raisebox{-18pt}{\includegraphics[height=0.12\textwidth, width = 0.12\textwidth]{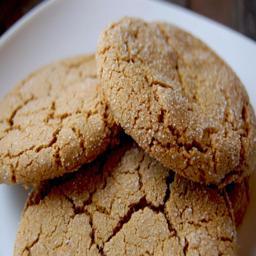}}&
 \raisebox{-18pt}{\includegraphics[height=0.12\textwidth, width = 0.12\textwidth]{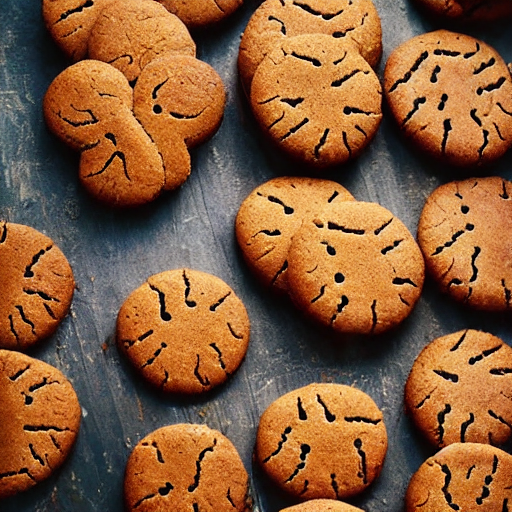}} &
 \raisebox{-18pt}{\includegraphics[height=0.12\textwidth, width = 0.12\textwidth]{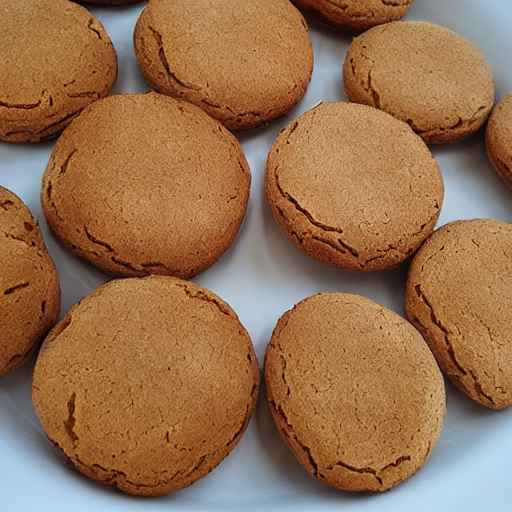}} &
 \raisebox{-18pt}{\includegraphics[height=0.12\textwidth, width = 0.12\textwidth]{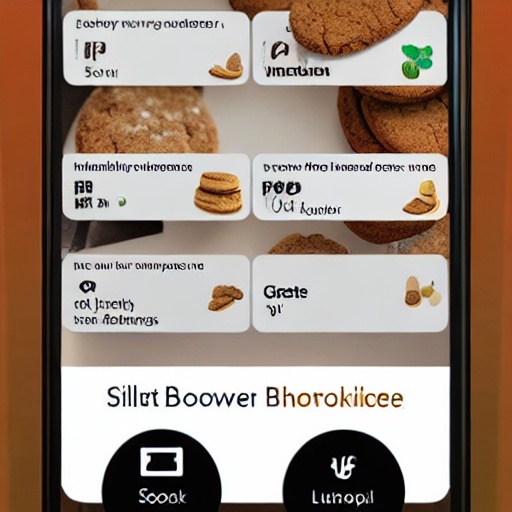}} &
 \raisebox{-18pt}{\includegraphics[height=0.12\textwidth, width = 0.12\textwidth]{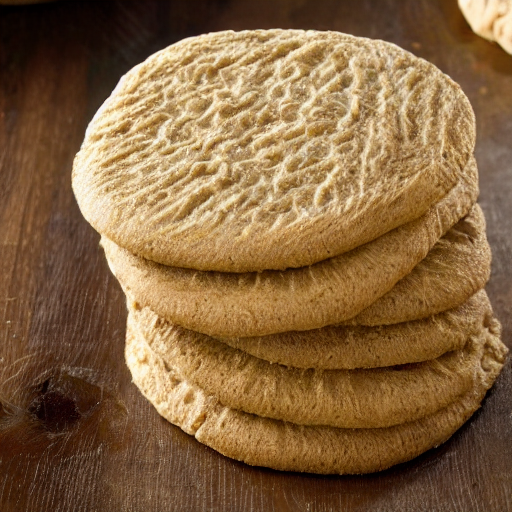}} &
 \raisebox{-18pt}{\includegraphics[height=0.12\textwidth, width = 0.12\textwidth]{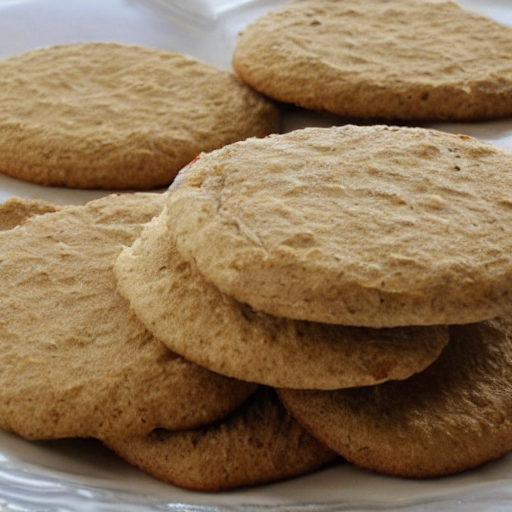}}&
 \raisebox{-18pt}{\includegraphics[height=0.12\textwidth, width = 0.12\textwidth]{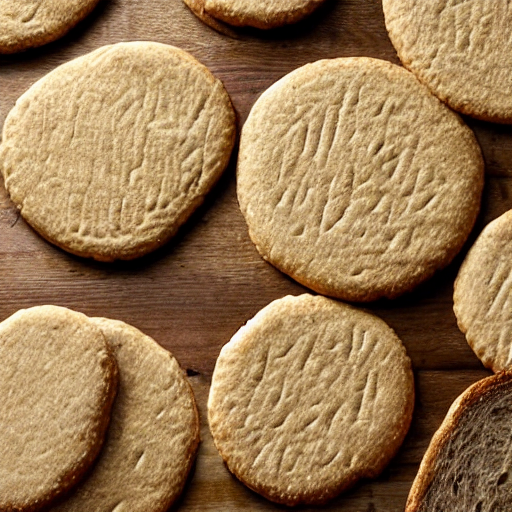}}\\
& \multicolumn{3}{m{0.36\textwidth+3\tabcolsep+\arrayrulewidth\relax}}{suffolkworkforce shared editor , agron app quotehomegrown  ginger biscuits recipe lovely } & \multicolumn{3}{m{0.36\textwidth+3\tabcolsep+\arrayrulewidth\relax}}{length bread wheat cookies round moist worthless perhaps}\\
 \midrule
  \raisebox{-18pt}{\includegraphics[height=0.12\textwidth, width = 0.12\textwidth]{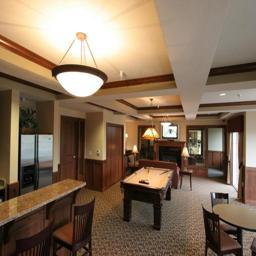}}&
 \raisebox{-18pt}{\includegraphics[height=0.12\textwidth, width = 0.12\textwidth]{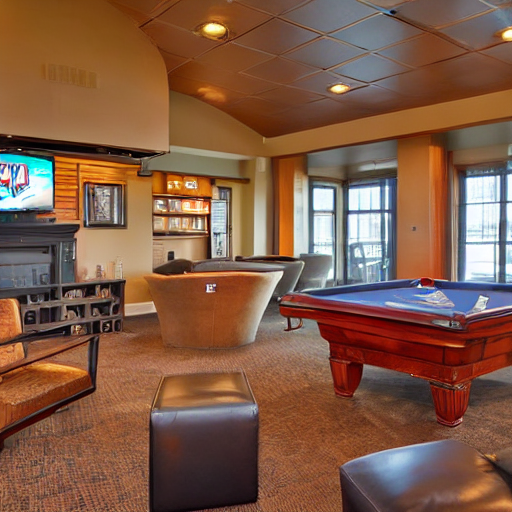}} &
 \raisebox{-18pt}{\includegraphics[height=0.12\textwidth, width = 0.12\textwidth]{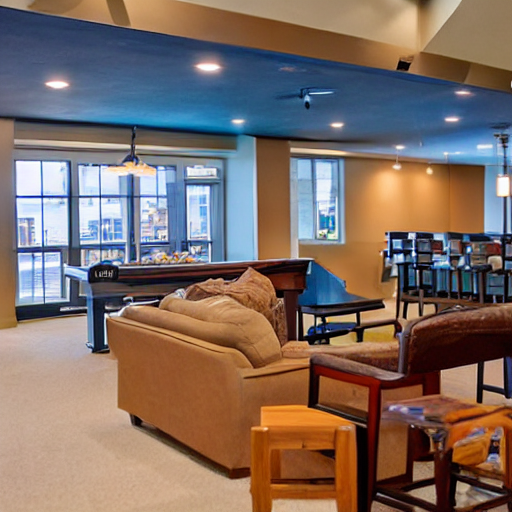}} &
 \raisebox{-18pt}{\includegraphics[height=0.12\textwidth, width = 0.12\textwidth]{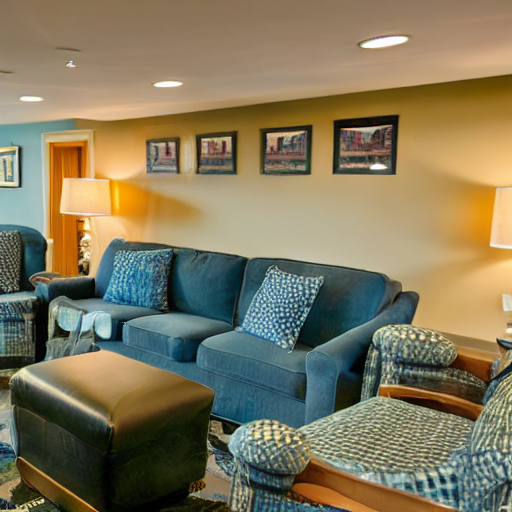}} &
 \raisebox{-18pt}{\includegraphics[height=0.12\textwidth, width = 0.12\textwidth]{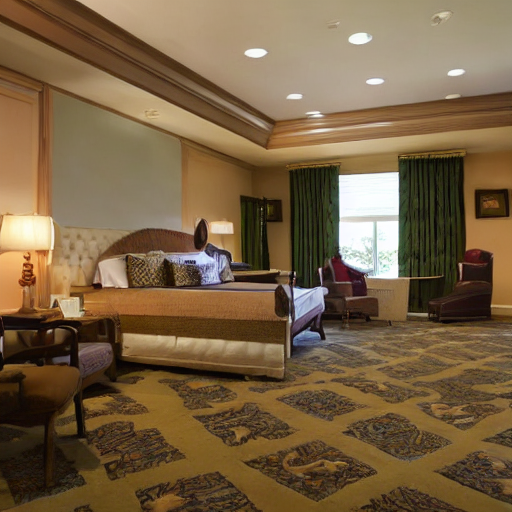}} &
 \raisebox{-18pt}{\includegraphics[height=0.12\textwidth, width = 0.12\textwidth]{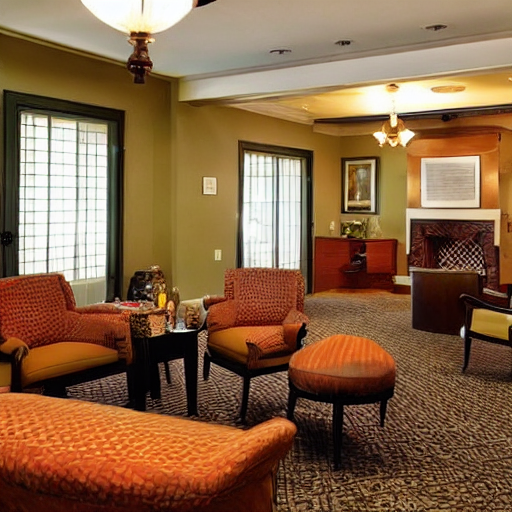}}&
 \raisebox{-18pt}{\includegraphics[height=0.12\textwidth, width = 0.12\textwidth]{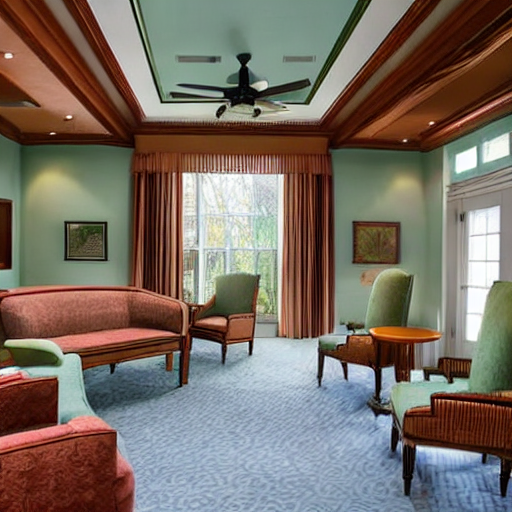}}\\
 & \multicolumn{3}{m{0.36\textwidth+3\tabcolsep+\arrayrulewidth\relax}}{renovated penthouse clubhouse ---- welch grand wisconwisindianfootball renovated ---- bluecntv occupoccup} & \multicolumn{3}{m{0.36\textwidth+3\tabcolsep+\arrayrulewidth\relax}}{plantation clubhouse proposed room wanting chamber renovated}\\
 \midrule
  \raisebox{-18pt}{\includegraphics[height=0.12\textwidth, width = 0.12\textwidth]{sec/figures/suppmat/6.jpg}}&
 \raisebox{-18pt}{\includegraphics[height=0.12\textwidth, width = 0.12\textwidth]{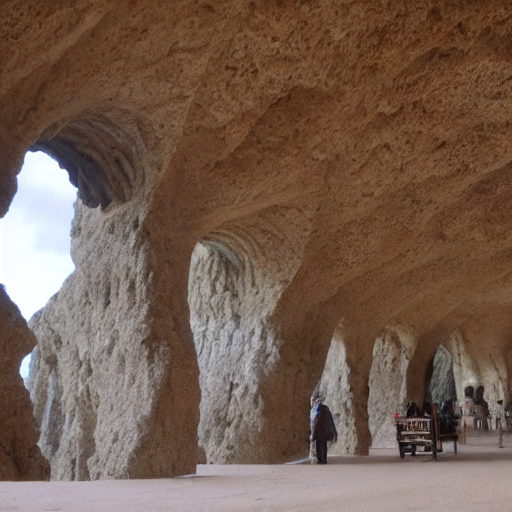}} &
 \raisebox{-18pt}{\includegraphics[height=0.12\textwidth, width = 0.12\textwidth]{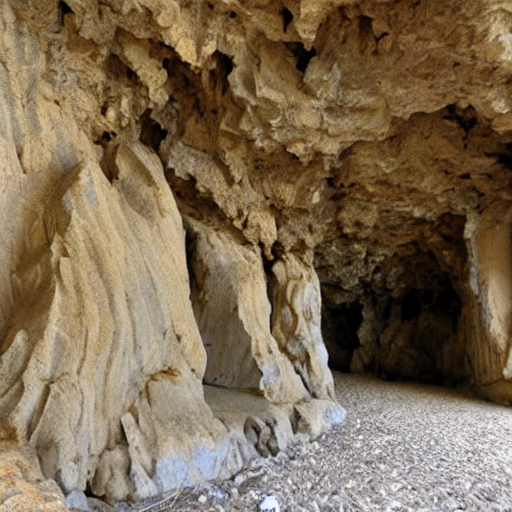}} &
 \raisebox{-18pt}{\includegraphics[height=0.12\textwidth, width = 0.12\textwidth]{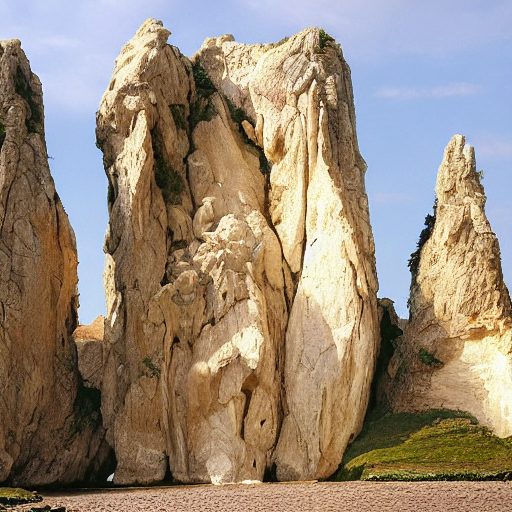}} &
 \raisebox{-18pt}{\includegraphics[height=0.12\textwidth, width = 0.12\textwidth]{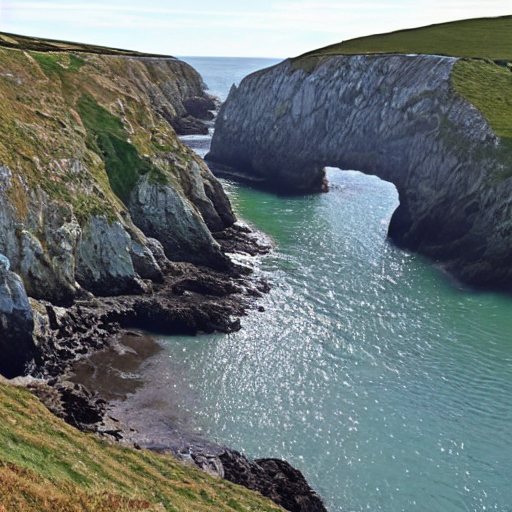}} &
 \raisebox{-18pt}{\includegraphics[height=0.12\textwidth, width = 0.12\textwidth]{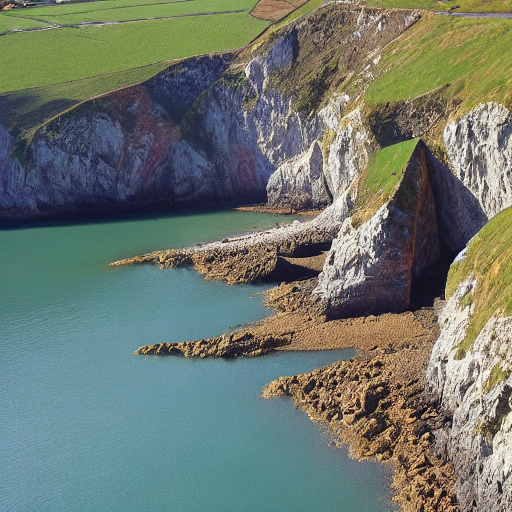}}&
 \raisebox{-18pt}{\includegraphics[height=0.12\textwidth, width = 0.12\textwidth]{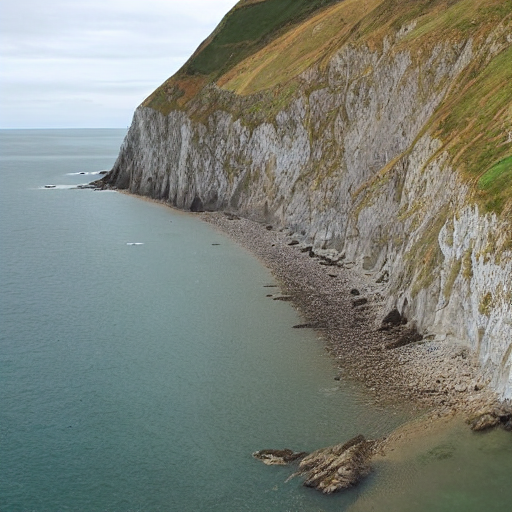}}\\
 & \multicolumn{3}{m{0.36\textwidth+3\tabcolsep+\arrayrulewidth\relax}}{discoverkeywords temps monasterdiscover caverirrational apostles chose marqutours gaubois september coastline } & \multicolumn{3}{m{0.36\textwidth+3\tabcolsep+\arrayrulewidth\relax}}{fitz caves conwy coast showing rt margaret vigilant margaret gilli cliffs}\\
  \midrule
  \raisebox{-18pt}{\includegraphics[height=0.12\textwidth, width = 0.12\textwidth]{sec/figures/image_similarity/15.jpg}}&
 \raisebox{-18pt}{\includegraphics[height=0.12\textwidth, width = 0.12\textwidth]{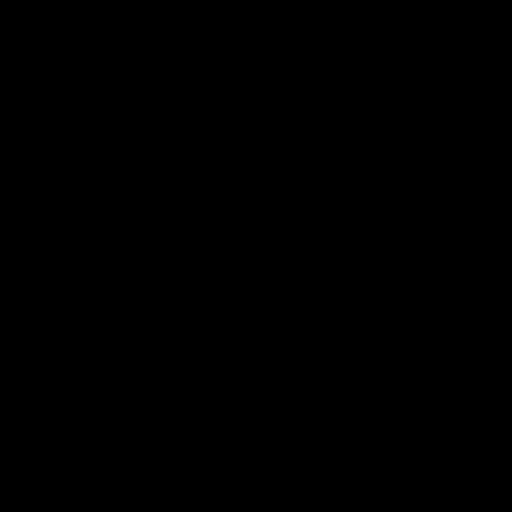}} &
\raisebox{-18pt}{\includegraphics[height=0.12\textwidth, width = 0.12\textwidth]{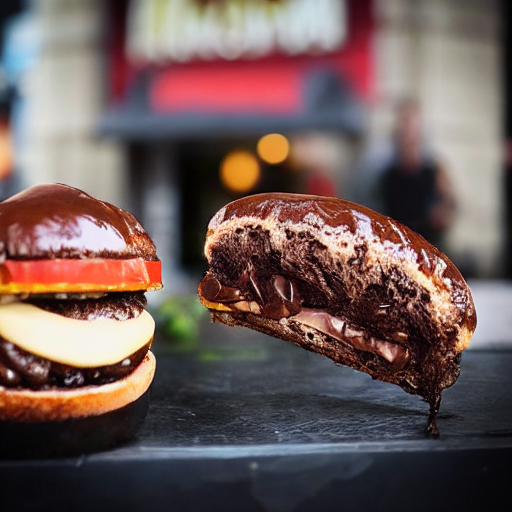}} &
 \raisebox{-18pt}{\includegraphics[height=0.12\textwidth, width = 0.12\textwidth]{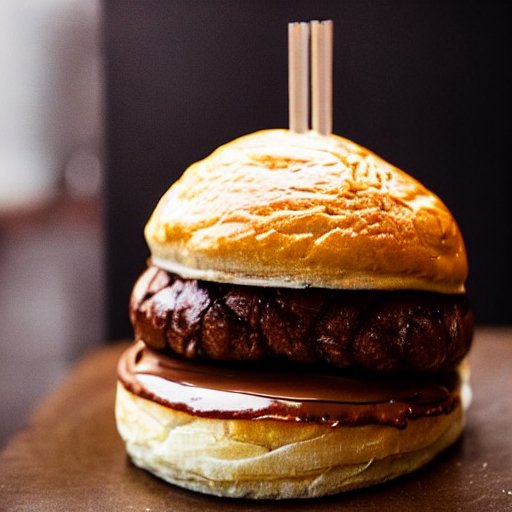}} &
 \raisebox{-18pt}{\includegraphics[height=0.12\textwidth, width = 0.12\textwidth]{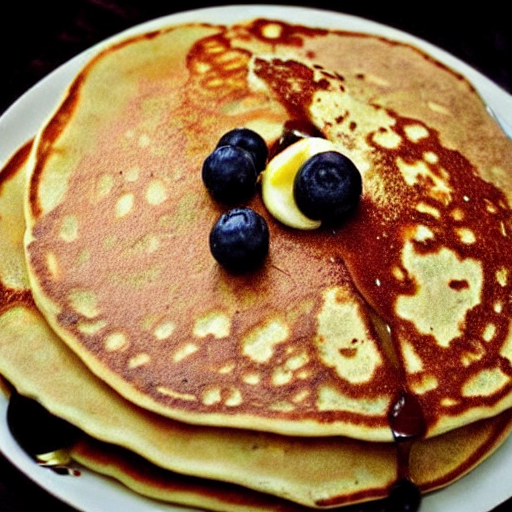}} &
 \raisebox{-18pt}{\includegraphics[height=0.12\textwidth, width = 0.12\textwidth]{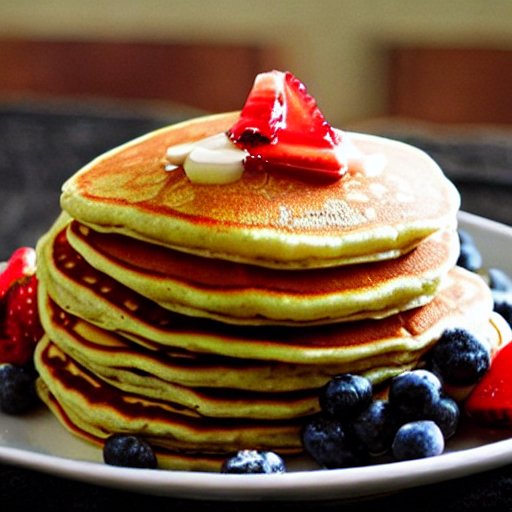}}&
 \raisebox{-18pt}{\includegraphics[height=0.12\textwidth, width = 0.12\textwidth]{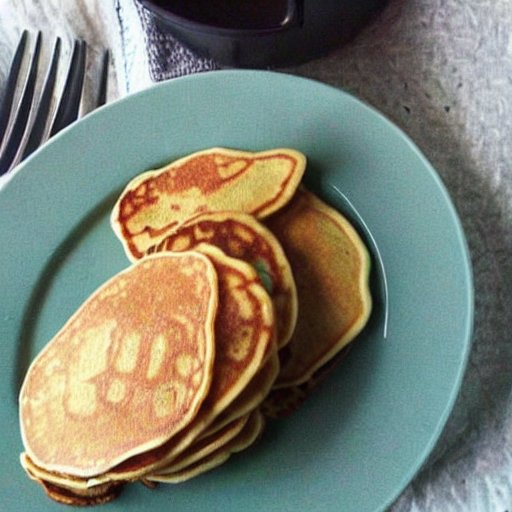}}\\
 & \multicolumn{3}{m{0.36\textwidth+3\tabcolsep+\arrayrulewidth\relax}}{nutella burger brioche consumed solves dis \img{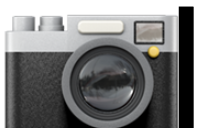}
 $|@$ wasting kx $[@$ vin johnny wimpoaching imagery } & \multicolumn{3}{m{0.36\textwidth+3\tabcolsep+\arrayrulewidth\relax}}{sometimes pancakes itis :) " petr soho}\\
\bottomrule
\end{tabularx}
\caption{{\bf Qualitative Comparison}. Target image, inverted prompts (from PEZ \cite{WenHPME23} and PH2P), and corresponding generated images.}
\label{tab:qualitative_PEZ_PH2P:supp}
\vspace{-5 mm}
\end{table*}
\begin{table*}[h]
 \centering
 \scriptsize
\begin{tabularx}{\textwidth}{@{}X*{3}{m{0.25\textwidth}}@{}}
\toprule
Target Image & {LDM+Adam} & {LDM+ all $t$} & {PH2P (Ours)}\\
\midrule
\raisebox{-32pt} {\includegraphics[height=0.12\textwidth, width = 0.12\textwidth]{sec/figures/ablations/21_000000001761.jpg}} & watched flying supremacy baptiste scho mueller at & airplanes sean moulin chuck claw peak intimidation  intimidation &  airplane photography skies myrtle striking farmer sometimes\\
\midrule
\raisebox{-30pt} {\includegraphics[height=0.12\textwidth, width = 0.12\textwidth]{sec/figures/ablations/67_000000006771.jpg}} & smokers concern criminal snippets geneva & rinking lexa cybermonday pilgrimage vendor apparently glimpse  pend ctar iterate &  visitors eyes hacked ' protesters reflects metoo\\
\midrule
\raisebox{-30pt} {\includegraphics[height=0.12\textwidth, width = 0.12\textwidth]{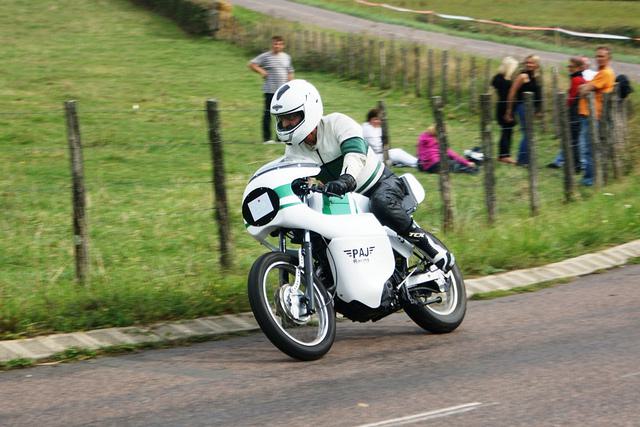}} & A : coventry motorcycles vina championship & sbk rbs triumph coventry   sprints slalom bhar classiccar ingu&  badgers motorcycle driving racers championship\\
\midrule
\raisebox{-30pt} {\includegraphics[height=0.12\textwidth, width = 0.12\textwidth]{sec/figures/ablations/83_000000007977.jpg}} & skater ffi actions fitt osc looking * & rotation " reversed skateboarding sitter skateboarding skill handshake  upside&   michal illusion precision skate sitter skateboarding routine\\
\midrule
\raisebox{-30pt} {\includegraphics[height=0.12\textwidth, width = 0.12\textwidth]{sec/figures/ablations/84_000000007991.jpg}} & :) dinner knows dayz examples vegetables dunno & shaped lower diameter saves tying carrots use &  carrots picked leftover wid mara lls munchies\\
\bottomrule
\end{tabularx}
\caption{{\bf Qualitative Comparison to Ablations.} 
Qualitative comparison of the prompts inverted and the diverse images generated with our PH2P to the baselines for ablation.}
\label{tab:ablations:supp}
\vspace{-5 mm}
\end{table*}
\begin{table*}
 \centering
 \scriptsize
\begin{tabularx}{\textwidth}{@{}*{9}{m{0.085\textwidth}}@{}}
\toprule
Target Image & \multicolumn{8}{l}{Attented regions in the target image for PH2P prompt inversion} \\
\midrule
  \raisebox{-8pt} {\includegraphics[height=0.10\textwidth, width = 0.10\textwidth]{sec/figures/attention/163_000000016439.jpg}} &
 \raisebox{-18pt} {\includegraphics[width = 0.10\textwidth,keepaspectratio]{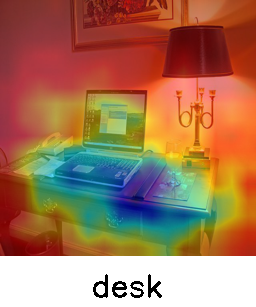}} &
\raisebox{-18pt}{\includegraphics[width = 0.10\textwidth,keepaspectratio]{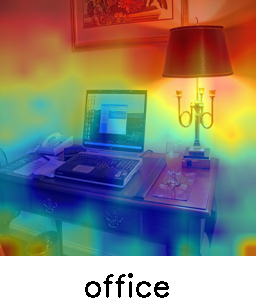}} &
\raisebox{-18pt}{\includegraphics[width = 0.10\textwidth,keepaspectratio]{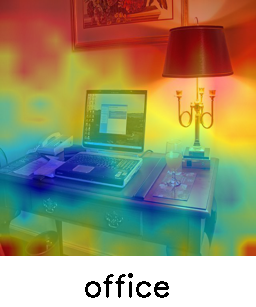}} &
\raisebox{-18pt}{\includegraphics[width = 0.10\textwidth,keepaspectratio]{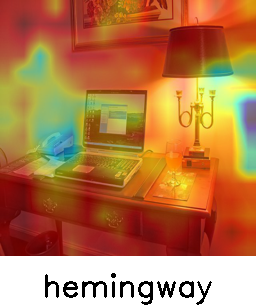}} &
\raisebox{-18pt}{\includegraphics[width = 0.10\textwidth,keepaspectratio]{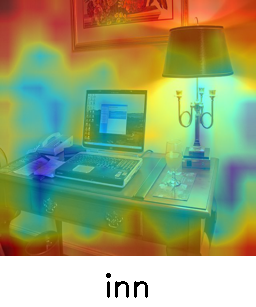}}&
\raisebox{-18pt}{\includegraphics[width = 0.10\textwidth,keepaspectratio]{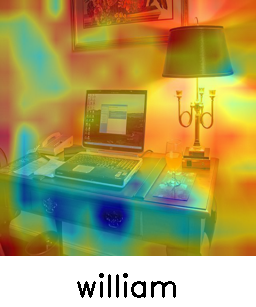}}
&
\raisebox{-18pt}{\includegraphics[width = 0.10\textwidth,keepaspectratio]{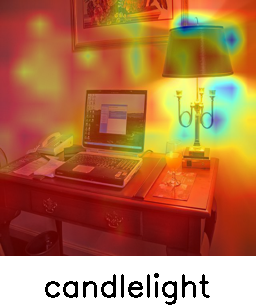}}
&
\raisebox{-18pt}{\includegraphics[width = 0.10\textwidth,keepaspectratio]{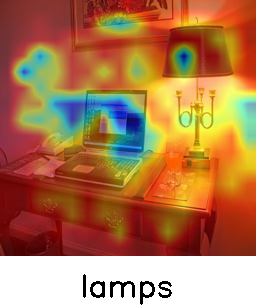}}
\\
\midrule
 \raisebox{-8pt} {\includegraphics[height=0.10\textwidth, width = 0.10\textwidth]{sec/figures/attention/417_000000047112_2.jpg}} &
 \raisebox{-18pt} {\includegraphics[ width = 0.10\textwidth,keepaspectratio]{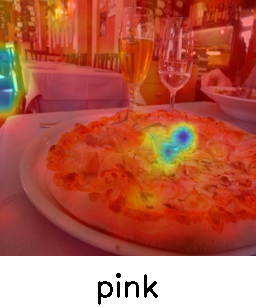}} &
\raisebox{-18pt}{\includegraphics[width = 0.10\textwidth,keepaspectratio]{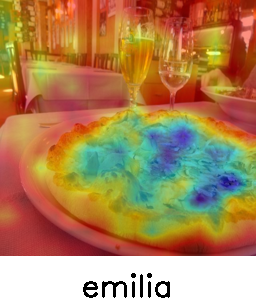}} &
\raisebox{-18pt}{\includegraphics[width = 0.10\textwidth,keepaspectratio]{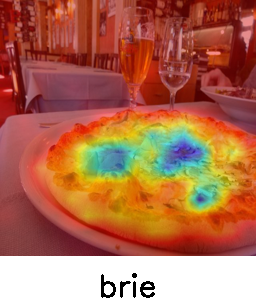}} &
\raisebox{-18pt}{\includegraphics[width = 0.10\textwidth,keepaspectratio]{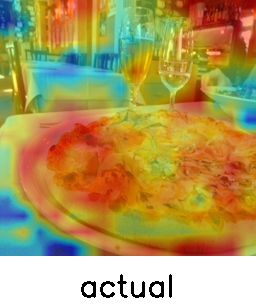}} &
\raisebox{-18pt}{\includegraphics[width = 0.10\textwidth,keepaspectratio]{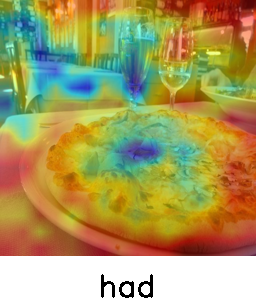}}&
\raisebox{-18pt}{\includegraphics[width = 0.10\textwidth,keepaspectratio]{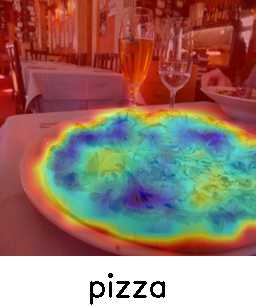}}
&
\raisebox{-18pt}{\includegraphics[width = 0.10\textwidth,keepaspectratio]{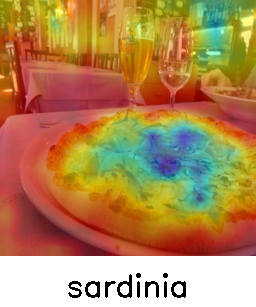}} &
\raisebox{-18pt}{\includegraphics[width = 0.10\textwidth,keepaspectratio]{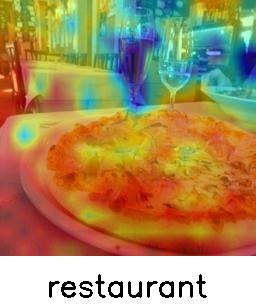}}
\\
\bottomrule
\end{tabularx}
\vspace{-2 mm}
\captionof{figure}{{\bf Application of Unsupervised Segmentation.} Additional results on illustration of the tokens and corresponding regions (that can be used for unsupervised segmentation; see \cite{abs-2303-11681}) obtained for the target images. Note the accuracy of both prompts and corresponding attention.} 
\label{tab:attention_maps:supp}
\vspace{-3 mm}
\end{table*}
\end{document}